\documentclass[letterpaper,10 pt,conference]{IEEEtran}
\IEEEoverridecommandlockouts

\usepackage{tikz}
\usepackage{amsmath}
\usepackage{amssymb}
\usepackage{booktabs}
\usepackage{comment}
\usepackage{color}
\usepackage{enumitem}
\usepackage{url}

\usepackage[ruled,vlined,linesnumbered]{algorithm2e}

\SetCommentSty{mycommfont}
\usepackage{algorithmic}
\usepackage{amsfonts}
\usepackage{caption}
\usepackage{graphicx}
\usepackage{multirow}
\usepackage{subcaption}
\usepackage{textcomp}
\usepackage[normalem]{ulem}
\usepackage{xspace}
\usepackage{verbatim}
\usepackage{authblk}

\ifCLASSOPTIONcompsoc
  \usepackage[%
    square,        
    comma,         
    numbers,       
    sort           
  ]{natbib}
\else
  \usepackage[%
    square,        
    comma,         
    numbers,       
    sort&compress 
  ]{natbib}
\fi

\newcommand{\change}[1]{\textcolor{black}{#1}}
\newcommand{\revise}[1]{\textcolor{black}{#1}} 
\newcommand{\eg}{\textit{e.g.}\xspace}
\newcommand{\ie}{\textit{i.e.}\xspace}
\newcommand{\etal}{\textit{et al.}\xspace}

\newcommand{\approach}{\textsc{DeepObliviate}\xspace}
\newtheorem{mydef}{Definition}

\newcounter{finding}

\newcounter{remark}
\newenvironment{remark}{
	\refstepcounter{remark}\par\smallskip
	\noindent \textbf{Remark:\xspace}\itshape
}
{
	\par\smallskip \normalfont
}

\begin{document}

	\date{}
	

	\title{\approach: A Powerful Charm for Erasing Data Residual Memory in Deep Neural Networks }
	
	\author[1,2]{\normalsize{Yingzhe He}}
	\author[1,2]{Guozhu Meng}
	\author[1,2]{Kai Chen}
	\author[1,2]{Jinwen He}
	\author[1,2]{Xingbo Hu}
	\affil[1]{\normalsize\emph{SKLOIS, Institute of Information Engineering, Chinese Academy of Sciences, China}}
	\affil[2]{\emph{School of Cyber Security, University of Chinese Academy of Sciences, China}} 
	
	\maketitle
	
	\thispagestyle{plain}

\begin{abstract}
Machine unlearning has great significance in guaranteeing model security and protecting user privacy. Additionally, many legal provisions clearly stipulate that users have the right to demand model providers to delete their own data from training set, that is, the right to be forgotten. 
The naive way of unlearning data is to retrain the model without it from scratch, which becomes extremely time and resource consuming at the modern scale of deep neural networks.  
Other unlearning approaches by refactoring model or training data struggle to gain a balance between overhead and model usability.

In this paper, we propose an approach, dubbed as \approach, to implement machine unlearning efficiently, without modifying the normal training mode. Our approach improves the original training process by storing intermediate models on the hard disk. Given a data point to unlearn, we first quantify its temporal residual memory left in stored models. The influenced models will be retrained and we decide when to terminate the retraining based on the trend of residual memory on-the-fly. Last, we stitch an unlearned model by combining the retrained models and uninfluenced models.
We extensively evaluate our approach on five datasets and deep learning models. 
Compared to the method of retraining from scratch, our approach can achieve 99.0\%, 95.0\%, 91.9\%, 96.7\%, 74.1\% accuracy rates and 66.7$\times$, 75.0$\times$, 33.3$\times$, 29.4$\times$, 13.7$\times$ speedups on the MNIST, SVHN, CIFAR-10, Purchase, and ImageNet datasets, respectively.
Compared to the state-of-the-art unlearning approach, we improve 5.8\% accuracy, 32.5$\times$ prediction speedup, and reach a comparable retrain speedup under identical settings on average on these datasets. 
Additionally, \approach can also pass the backdoor-based unlearning verification.

\end{abstract}

	\section{Introduction}\label{sec:intro}

In recent years, deep learning has gained extensive progress in image classification, speech recognition, natural language processing, and etc. 
To handle more complex tasks and variable decision-making scenarios, deep learning models (DLMs) have evolved from LeNet~\cite{u13-Lenet} of simple network structures to AlexNet~\cite{u14-Alexnet}, ResNet~\cite{u16-resnet}, VGGNet~\cite{u15-vgg}, GoogLeNet~\cite{u17-Googlenet} and other deeper models. 
Additionally, a massive amount of high quality data is required and collected by model providers from multitudinous data providers. 
With the increasing demand of privacy and security, there emerges a new requirement of erasing trained data, dubbed as machine unlearning conventionally, from a well-trained model~\cite{u1-DBLP:conf/sp/CaoY15}.

\change{In a deep learning task, data providers offer their data and model providers collect the data to train DLMs.} 
The requirements for unlearning data come from both two sources.
The first \change{and more important} requirement comes from data providers, who want their offered data removed from the model to prevent privacy leak and abuse. 
DLMs may learn private information from the training data. For instance, an image of house number likely exposes its owner's \change{home address, building material, housing color and other information}~\cite{u18-SVHN}. 
A medical record that elaborates a patient's medical history can be revealed from a membership inference attack~\cite{mia}. \change{Attackers can infer whether a user purchased a product based on shopping records~\cite{u19-purchase}.}
More importantly, data owners have the legal right of \change{removing their private data from the trained model, that is, \textit{the right to be forgotten}.}
There are many bills to guarantee this right such as the General Data Protection Regulation (GDPR)~\cite{GDPR}, the California Consumer Privacy Act (CCPA)~\cite{CCPA} and Amended Act on the Protection of Personal Information (APPI)~\cite{APPI}. \change{They incur} a mandatory legal obligation of model providers to unlearn data.

The unlearning requirements from model providers \change{include removing polluted or outdated data.} 
On one hand, DLMs are suffering from poisoning attacks~\cite{9-DBLP:conf/sp/JagielskiOBLNL18, 133-DBLP:journals/corr/abs-1808-08994, 116-DBLP:journals/corr/abs-1804-00792}, where training data is polluted by crafted attacking data samples. 
Poisonous samples can undermine the usability of DLMs~\cite{166-DBLP:conf/ecai/XiaoXE12, 9-DBLP:conf/sp/JagielskiOBLNL18}, and implant a backdoor into the model~\cite{116-DBLP:journals/corr/abs-1804-00792}. 
Consequently, model providers need to remove poisonous samples as well as the caused influences to model.
On the other hand, some training data is time-sensitive and will become out-of-date or even wrong in future, \ie, concept drift~\cite{concept-drift-1}. It may degrade the performance of DLMs~\cite{concept-drift-2}. Model providers have to eliminate the influence imposed by the outdated data for usability.

Attributed to above requirements, the technology of machine unlearning (hereafter we use deep unlearning to represent the unlearning techniques towards DLMs) appears and increasingly gains researchers and practitioners' concern. 
Conceptually, a deep unlearning process can be interpreted as eliminating the influence to model of data points requested for unlearning. \change{If the normal learning performs an ``addition'' operation, then unlearning performs a ``subtraction'' operation to the model.} 
A naive unlearning (baseline) can be accomplished by \change{removing the unlearned data}, and retraining the remaining data points from scratch~\cite{u3-DBLP:journals/corr/abs-1912-03817,u2-DBLP:conf/nips/GinartGVZ19, u10-DBLP:conf/eurocrypt/GargGV20}. 
As we all know, it is never trivial to train a DLM at the modern scale of dataset and model. 
Under such circumstances, the naive unlearning undoubtedly requires \change{a lot of computing resources and retraining time.} 
Frequent unlearning is even unaffordable in reality. 
Therefore, it motivates researchers to develop fast yet cost-effective approaches of deep unlearning.

Prior studies for improving the efficiency of machine or deep unlearning roughly fall into two categories: \emph{parameters manipulation}, that is to directly update model parameters to offset the impact of deleted data on the model~\cite{u4-DBLP:journals/corr/abs-1911-03030, u5-DBLP:journals/corr/abs-2002-02730, u6-DBLP:journals/corr/abs-2003-10933}, and; \emph{dataset reorganizing}, where model providers reorganize the training data and train one or several new models, and these models work collaboratively for a consensus prediction~\cite{u1-DBLP:conf/sp/CaoY15, u2-DBLP:conf/nips/GinartGVZ19, u3-DBLP:journals/corr/abs-1912-03817}.
Parameters manipulation is effective in machine learning models like K-means~\cite{u21-K-means}, decision tree~\cite{u22-decision-tree}, and SVM~\cite{u23-SVM}. 
The approach of model sharding (SISA)~\cite{u3-DBLP:journals/corr/abs-1912-03817} can apply to DLMs, but it loses much model accuracy with the increase of shard number. Liu~\etal~\cite{u6-DBLP:journals/corr/abs-2003-10933} also propose an unlearning method on DLMs, but it only works in federated learning.

\begin{figure}
	\centering
	\includegraphics[width=0.45\textwidth]{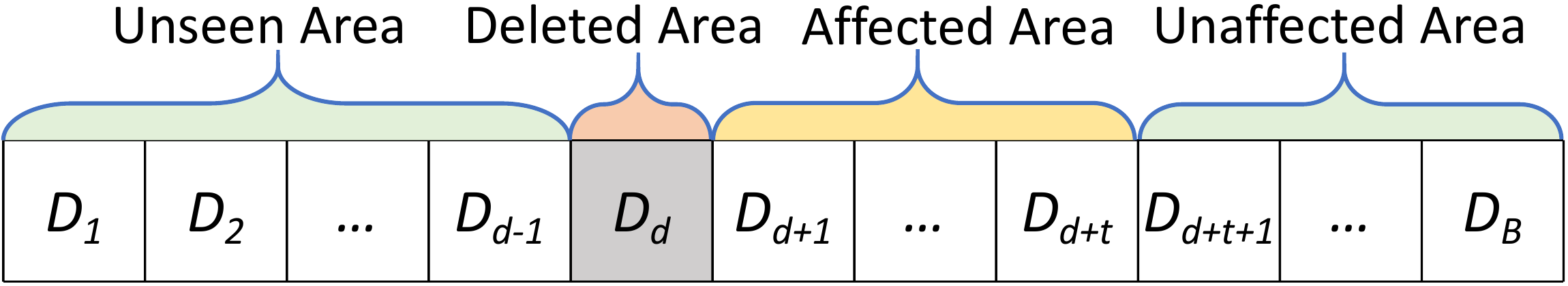}
	\caption{\change{An illustration example for unlearning in \approach, where one data block $D_i$ is requested to be deleted from the training dataset $\{D_1, \dots, D_B\}$.}}\label{fig:model-stitch-idea} 
\end{figure}

To this end, we make the first attempt to investigate and quantify the influence to model parameters of unlearned data, termed as \emph{temporal residual memory}, in an iterative training process. 
Through an empirical study, we observe that the temporal residual memory is subject to exponential decay which fades at an increasing rate over time (see Section~\ref{sec:problem}).
Based on this phenomenon, we develop a \approach approach\footnote{``Obliviate'' is a powerful charm to wipe out human's memory used by Hermione Granger in J. K. Rowling's Harry Potter series. Our \approach aims to effectively unlearn data from deep neural networks.} for a fast yet cost-effective deep unlearning.  
Our approach can offset the impact of unlearning data on the model, reduce retraining overhead efficiently, and make no significant changes to the original model without introducing additional security risks and maintenance cost.
More specifically, we retain the intermediate models for training each \revise{block} (detailed definition in Section~\ref{sec:over:symbo}), and divide them into four areas as per temporal residual memory, \change{as shown in Figure~\ref{fig:model-stitch-idea}.} 
The first \emph{unseen} area contains all the models before the arrival of unlearned data. The second \emph{deleted} area contains the unlearned data. The third \emph{affected} area covers the models with prominent residual memory which need to be retrained, and the fourth \emph{unaffected} area is where the residual memory extinguishes. 
To determine the affected area, we introduce the parameter change vector to measure the residual memory. We adopt the detrended fluctuation analysis~\cite{DFA-1994} to calculate when this memory can be ignored and to terminate our retraining. As a result, an unlearned model can be stitched by reusing unseen and unaffected models, and retraining the affected areas.

Our approach is extensively evaluated on multiple datasets, including MNIST~\cite{mnist}, SVHN~\cite{u18-SVHN}, CIFAR-10~\cite{cifar10}, Purchase~\cite{u19-purchase}, and ImageNet~\cite{imagenet}, and a variety of DLMs such as LeNet~\cite{u13-Lenet}, ResNet~\cite{u16-resnet}, VGG~\cite{u15-vgg}. \change{When unlearning from 1 data point to 1\% data of training set, results show that \approach can reduce over 98$\sim$95, 98$\sim$82, 97$\sim$85, 96$\sim$94, 92$\sim$86 (\%) retraining costs, and achieve 99.0$\sim$96.9, 95.0$\sim$93.1, 91.9$\sim$90.8, 96.7$\sim$95.7, 74.1$\sim$68.7 (\%) accuracy on these five datasets (top-1 accuracy for ImageNet), respectively.} Under the identical experiment setting, \approach outperforms SISA~\cite{u3-DBLP:journals/corr/abs-1912-03817} by an improvement of \change{5.8\% accuracy, 1.01$\times$ retrain speedup, and 32.5$\times$ prediction speedup under the same storage overhead on average on these datasets.}
In addition, our approach can apply to varying unlearning scenarios and achieve superior results to date in deep unlearning. 
It takes only 19\% of naive unlearning efforts for multiple data deletions, and supports the operation of data deletion at any time. 
\change{We also adopt the backdoor-based verification~\cite{u7-DBLP:journals/corr/abs-2003-04247} to guarantee that our method has already unlearned the data.}

\noindent\textbf{Contributions}. We make the following contributions.
\begin{itemize}[leftmargin=*]
    \item We make the first attempt to quantify the temporal residual memory of unlearned data in a gradient-based training. We measure this residual memory through the difference vector between parameter change vectors on two models. It reflects how training data impacts on model parameters, and how this influence changes over time. 
    \item We develop a simple, fast and cost effective approach, \approach, that makes no significant changes to original models, requires less additional computation, and can handle multiple unlearning requests such as single, bulk, and frequent deletions. 
    \item We conduct an extensive evaluation with five deep learning models on five datasets. \approach provides a superior ability of unlearning data compared to the-state-of-the-art in the same setting. Additionally, we implement a backdoor-based unlearning verification that further proves the successfulness of deep unlearning.
 \end{itemize}

\section{Background}\label{sec:background}

In this section, we give a brief introduction of machine unlearning and four evaluation criteria.

\subsection{Machine Unlearning}

Different from machine learning that builds a mathematical model from data, machine unlearning can be regarded as a reverse process that drops data from the model. 
It is one of bionic technologies to imitate the function of the human brain. 
Humans have an efficient way to unlearn memory, especially unpleasant memory~\cite{machineunlearning}, as the experience that is seldom or never recalled is more likely to be forgotten. 
It used to be a defect for artificial neural networks since forgetting the previously seen information can lead to ``catastrophic interference''~\cite{catastrophic,catastrophic2}. 
The defect of amnesia in neural networks can be countered by technologies like latent learning~\cite{latentlearning} and self-refreshing memory~\cite{refreshing}. 
However, in this study, machine unlearning is one type of ability of DNNs. 
It is one subjective and intentional behavior of model providers to eliminate all the possible effects brought by specific data on the model.

Since the training data undergoes a number of complicated and even non-deterministic transformations before shaping a model, it is not easy at all to accurately measure the effects of part of data. 
The difficulty is further increased in the scenario of DNNs. Cook and Weisberg~\cite{inf,inf2} first propose \emph{influence functions} to approximate the leave-one-out cross-validation estimation of prediction variance. 
Koh and Liang use an influence function to correlate training data with the corresponding predictions~\cite{u27-influence2017icml}. 
To the best of our knowledge, there is no quantitative analysis yet of the influence exerted by training data to the model in a machine unlearning task.

\subsection{Formalization of Machine Unlearning}

Given a training process, we assume $D: \{x_1, x_2, \ldots, x_n\}$ as the training data, $F$ as a specific learning algorithm, and $M$ as the trained model. As such, we have $F: D \rightarrow M$, or $F(D) = M$ for simplicity, which means the model $M$ is built with the learning algorithm $F$ on dataset $D$. A deep unlearning operation on deleted $x_d$ ($x_d \in D$) can be represented as $U(x_d; M)$. Let $M'$ be the unlearned model, and we get $M' = U(x_d; M)$. 

\begin{mydef}\label{def:naive}
	\textbf{(Naive Unlearning)} \revise{Supposing} the model never sees the data point $x_d$, and thereby performs the retraining on the remaining $n-1$ data points, denoted as $D\backslash \{ x_d\}$. In this manner, the unlearned model is computed as $M' = F(D\backslash \{ x_d\})$, and we regard it as a naive unlearning. 
\end{mydef}

Without the consideration of computing and maintaining costs, naive unlearning is the perfect method for data deletion. 
However in reality, naive unlearning undoubtedly requires a large amount of computing time and resources far beyond the budgets of the majority of model providers. 
Alternatively, researchers start to employ other unlearning methods to circumvent this difficulty.
For instance, Bourtoule \etal propose a model sharding approach to partition the original model on the big dataset into several models on small datasets to reduce the retraining efforts~\cite{u3-DBLP:journals/corr/abs-1912-03817}. Liu \etal record the parameters during training and infer the changes for an unlearning operation in federated learning~\cite{u6-DBLP:journals/corr/abs-2003-10933}.
These unlearning approaches are essentially an optimized approximation to naive unlearning that will certainly produce either slight or significant errors compared to the ideal. 

To fairly and accurately evaluate unlearning approaches, we propose four criteria as well as their formal definitions. 
It is worthy mentioning that they are not completely orthogonal to the six goals in \cite{u3-DBLP:journals/corr/abs-1912-03817}, but these four criteria can be well quantified and measured. 
As aforementioned, $M'_n: F(D\backslash \{x_d\})$ is the naive unlearning, and $M'_u : U(x_d; F(D))$ is the proposed unlearning approach. We assume there are a set of samples $X$ for evaluating, and their true labels are $Y: \{y_1, y_2, \ldots, y_n\}$. 
$Y_n : \{y^n_1, y^n_2, \ldots, y^n_n\}$ and $Y_u : \{y^u_1, y^u_2, \ldots, y^u_n\} $ are the predicted labels by naive unlearning $M'_n(X)$ and proposed unlearning $M'_u(X)$, respectively.

\begin{itemize}[leftmargin=*]
	\item \textbf{Consistency}. It denotes how similarly the two models behave in front of same test samples. Given the same test samples, consistency measures how many samples the two models predict the same result. For an unlearning approach, this metric quantifies its gap with the naive unlearning. Consistency is computed as $|\{(y^n_i, y^u_i)~ |~ y^n_i = y^u_i \land 1 \leq i \leq n\}| / | Y |$. 

	\item \textbf{Accuracy}. Literally, it means how likely the model can correctly predict test samples. It reveals the usability of a model, and the model with low accuracy is useless in reality. That implies if the unlearning process undermine the accuracy of the original model, it is ineffective. 
	Accuracy can be computed as $|\{(y^u_i, y_i)~ |~ y^u_i = y_i \land 1 \leq i \leq n\}| / | Y |$ for an unlearning approach. 
 
	\item \textbf{Unlearning cost}. To fulfill unlearning target, researchers strive to reduce retraining time. The prediction overhead should also be considered when providing service. Additionally, model providers may spend storage cost for storing temporary data. In this study, we consider the time cost of retraining and prediction, and the storage cost in total. 
	
	\item \textbf{Verifiability}. It is demanded from users to verify whether or not the model provider has successfully unlearned the requested data. \change{A verification function $f$ can make a distinguishable check by $f(F(D), U(x_d; F(D))) \Leftrightarrow bool$ after unlearning $x_d$. Taking backdoor-based verification~\cite{u7-DBLP:journals/corr/abs-2003-04247} for example, if the pre-injected backdoor by unlearned $x_d$ is verified as being existing in $F(D)$ but removed from $U(x_d; F(D))$, the unlearning method $U$ is verified. }
\end{itemize}

	\section{Overview}\label{sec:new-overview}

\revise{In this section, we first introduce our approach from a whole perspective, and then we propose the temporal residual memory used in our unlearning approach.}

\begin{figure*}[t]
	\begin{subfigure}{.32\textwidth}
		\centering
		\includegraphics[width=\linewidth]{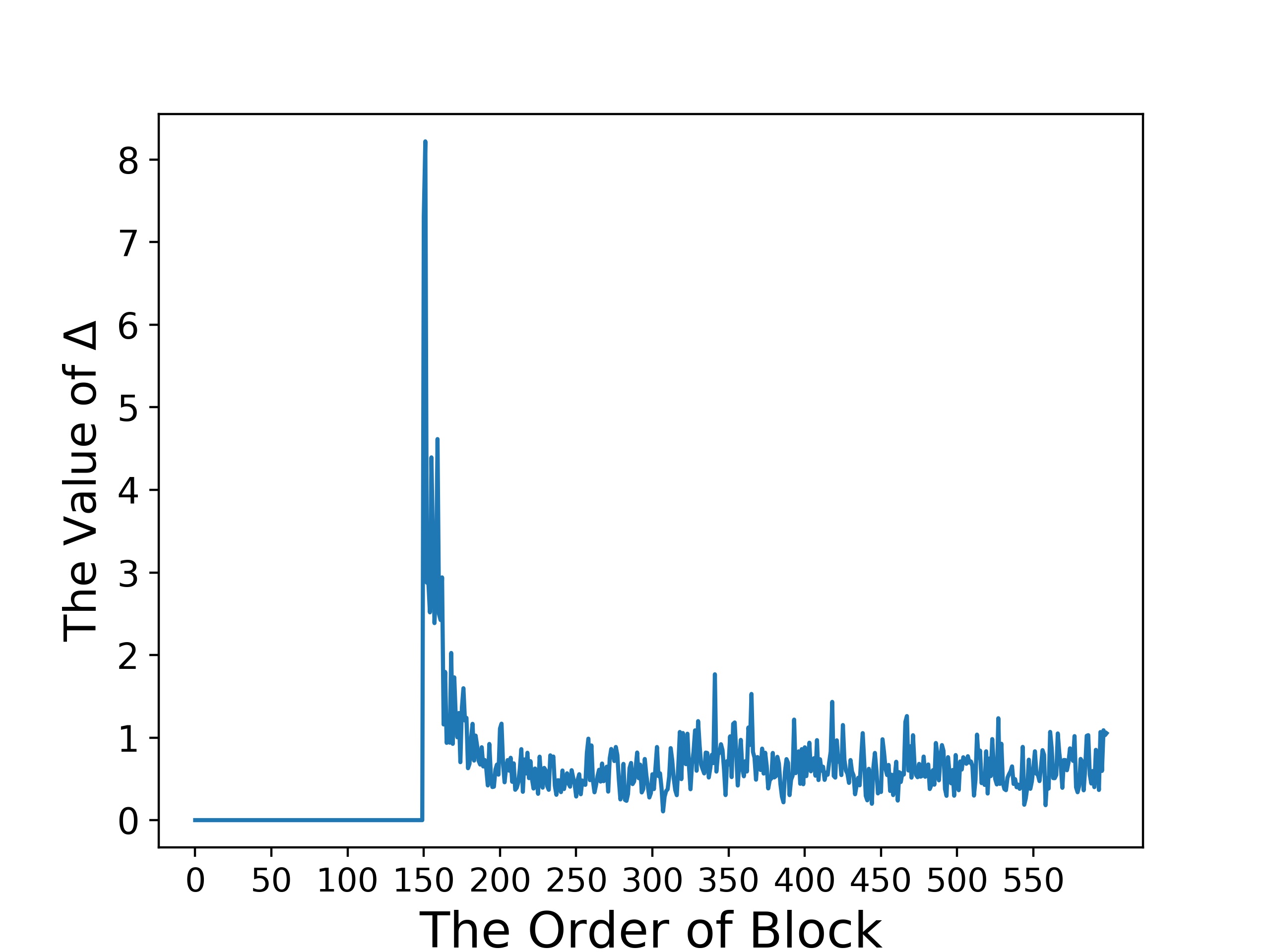}  
		\caption{Change curve of $\Delta$ over time under $B=600$. \revise{The deleted data locates in the 150th block.}}\label{fig:curve100}
	\end{subfigure}
	\hfill
	\begin{subfigure}{.32\textwidth}
		\centering
		\includegraphics[width=\linewidth]{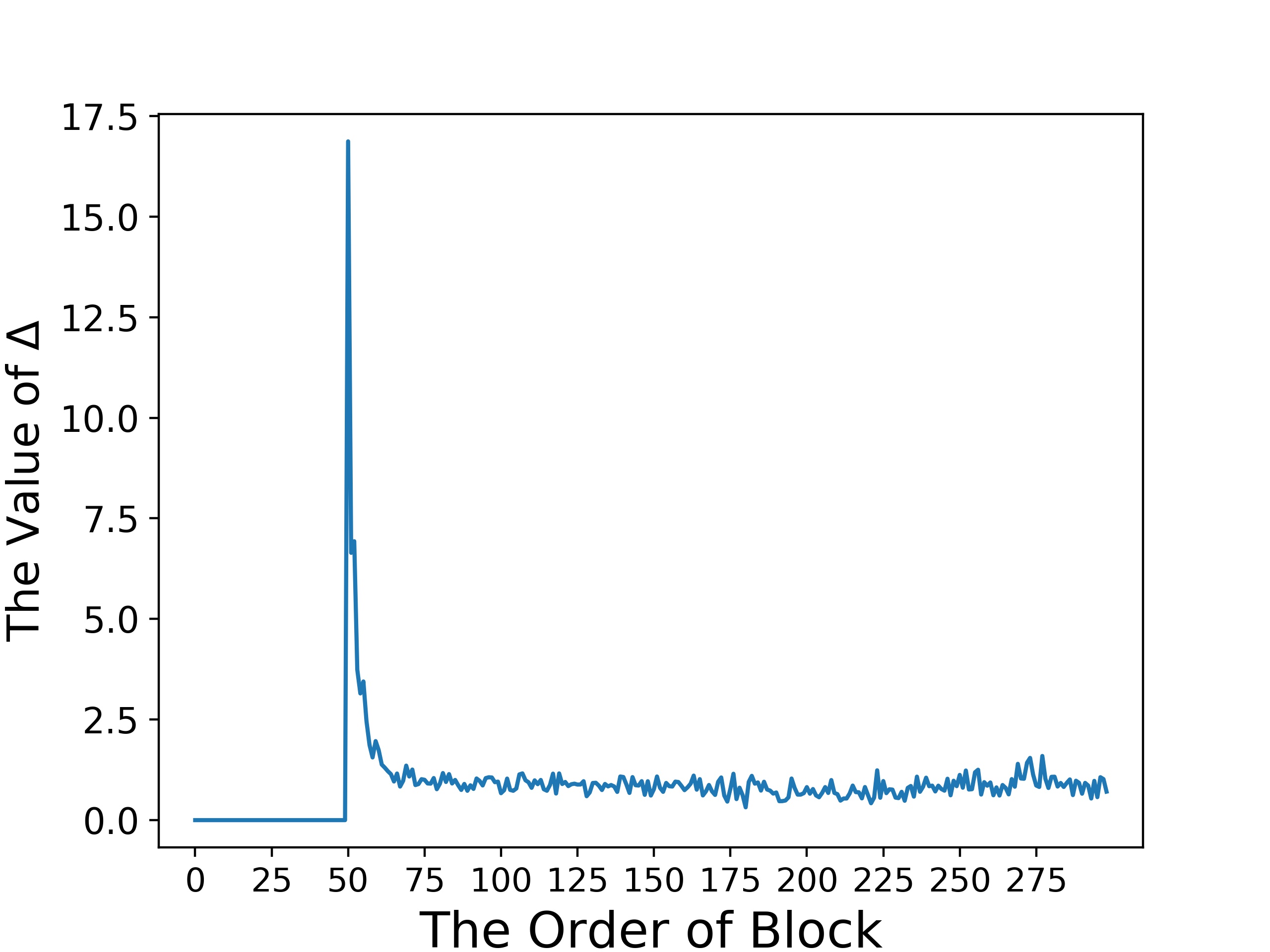}  
		\caption{Change curve of $\Delta$ over time under $B=300$. \revise{The deleted data locates in the 50th block.}}\label{fig:curve200}
	\end{subfigure}
	\hfill
	\begin{subfigure}{.32\textwidth}
		\centering
		\includegraphics[width=\linewidth]{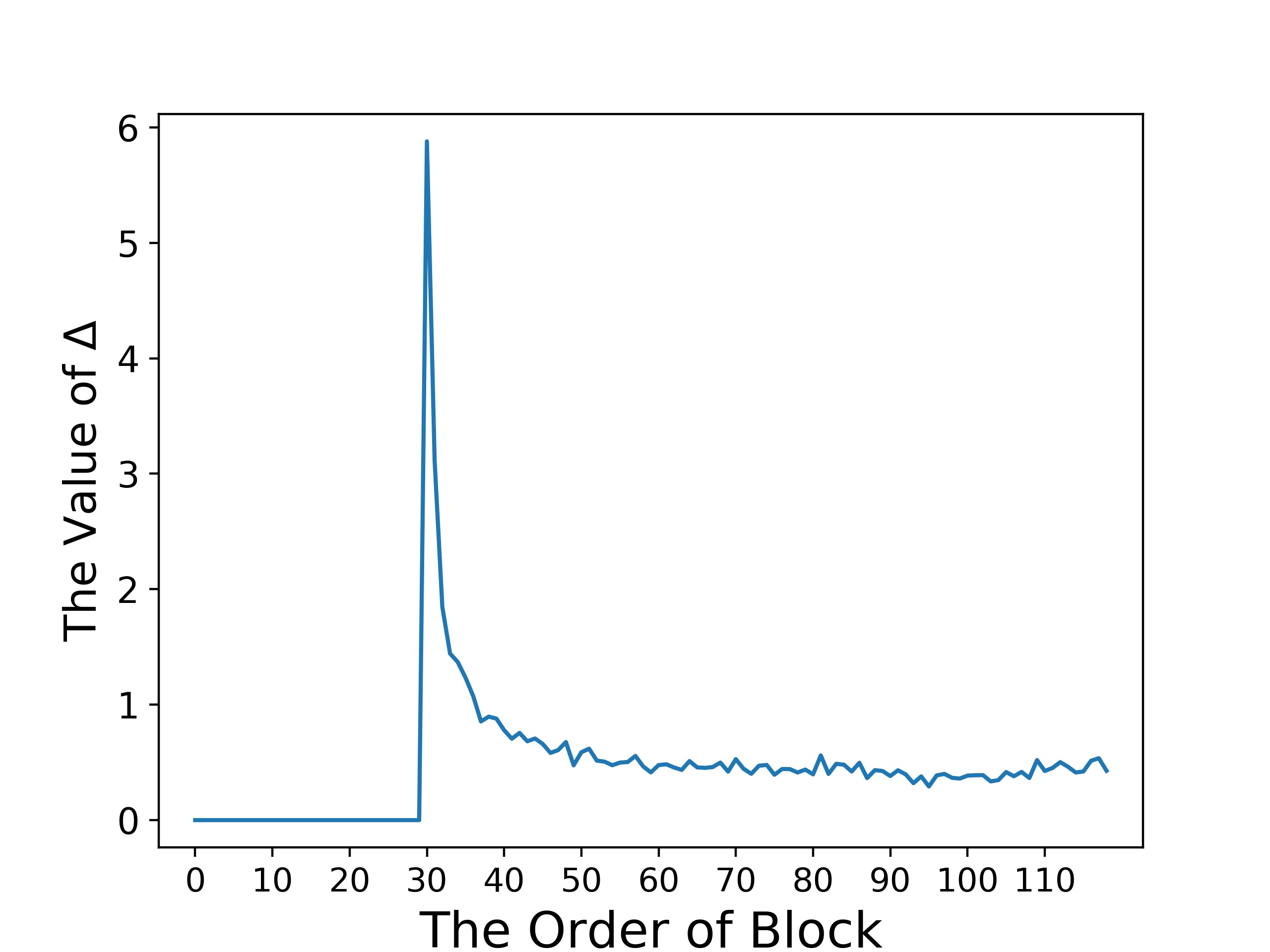}  
		\caption{Change curve of $\Delta$ over time under $B=120$. \revise{The deleted data locates in the 30th block.}}\label{fig:curve500}
	\end{subfigure}
	
	\caption{Residual memory left by unlearned data in three settings}
	\label{fig:residual}
\end{figure*}

\subsection{Workflow and Symbolization} \label{sec:over:symbo}

\revise{In our approach, the entire dataset $D$ is totally divided into some disjoint small datasets. Each small dataset is called \textit{a block}. When we divide blocks, it is best to ensure that the number of data points with the same label is uniform in each block. We use $B$ to represent the number of blocks, and $D=\{D_1, \cdots, D_B\}$ ($D_i$ is the $i$-th block). $\bigcap_i D_i=\emptyset$ and $\bigcup_i D_i=D$, for $i\in \{1, \cdots, B\}$. Each block contains $|D|/B$ data (some blocks contain $|D|/B+1$ data if it is not divisible).}

\revise{The training process takes each block as a complete dataset for multi-epoch training, and then trains the next block. After training each block, we store the model parameters at this time ($M_i$ for $i$-th block). When the entire dataset $D$ has been trained, we will obtain $\mathit{B}$ stored models, that is $\{M_1, \cdots, M_B\}$.}

\revise{Our next step is to use these stored models to carry out retraining in an unlearning request. First, we consider where a retraining is started. Supposing the unlearned data point is located in the $d$-th block, so it has no influence on the previous $d-1$ blocks and stored models. We just need to start the retraining process from the model $M_{d-1}$, and train the $d$-th block without the unlearned data point. This part is easy to understand and reduces the retraining overhead.}

\revise{Second, we consider where a retraining is terminated, which is also the focus of our approach. The naive idea of retraining until the last block is a waste of time. Through our experiments, we find that when the model is retrained at the $(d+t)$-th block ($t>0$), the unlearned data point will not have new influence on subsequent training (not that there is no influence, but the influence is very small and basically unchanged). This means that the influence of unlearned data on model $M_{d+t}$ is almost the same as it on model $M_B$. Directly subtracting the two models will cancel out the influence. So we terminate the retraining at the $(d+t)$-th block, and ``stitch'' the subtractive model to the retrained model. The 
entire unlearning process of \approach can be briefly understood as:}
\begin{equation}\label{equ:whole-process}
\begin{split}
    M' = Train(\{D'_d, D_{d+1}, \cdots, D_{d+t}\}|M_{d-1}) \oplus  (M_B \ominus  M_{d+t})
\end{split}
\end{equation}
\revise{Where $M'$ is the unlearned model. $D'_d$ deletes the unlearned data from $D_d$. $Train(u|v)$ means training the dataset $u$ based on the initial model $v$. $\oplus$ and $\ominus$ implement addition and subtraction operations on models.
More details of \approach and how to calculate $t$ will be expanded in Section~\ref{sec:problem} and Section~\ref{sec:approach}.}

\begin{figure*}
	\centering
	\includegraphics[width=0.8\textwidth]{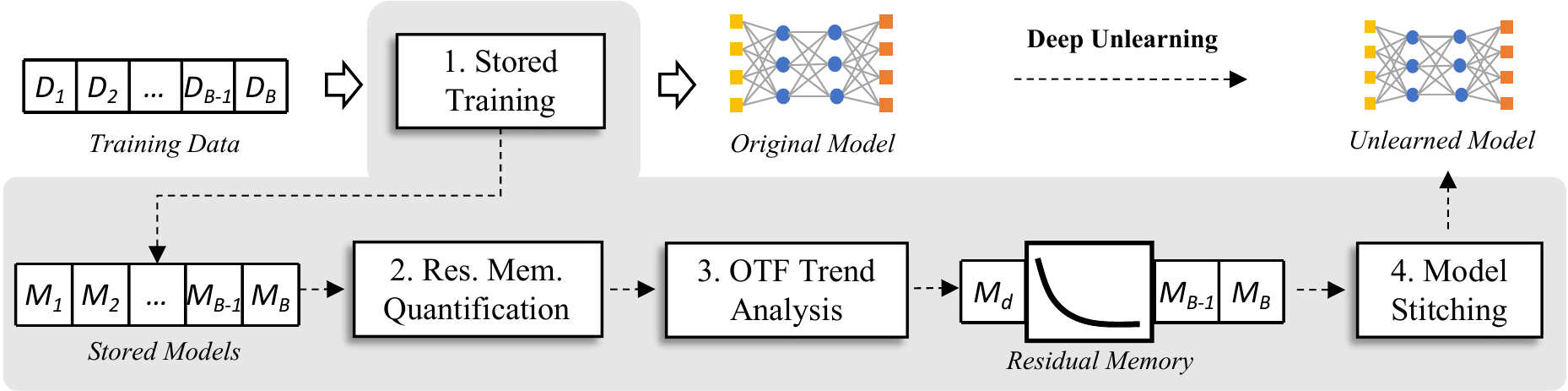}
	\caption{System overview of \approach}\label{fig:overview}
\end{figure*}
\subsection{Temporal Residual Memory in Unlearning}\label{sec:problem}

It is well known that gradient-based neural networks suffer from the vanishing gradient problem~\cite{vanishing-gradient} with the increasing number of layers. This makes the deeper network layers apt to forgetting the input data. That is why Long Short-Term Memory (LSTM)~\cite{u24-lstm} and Residual Networks (ResNet)~\cite{u16-resnet} are invented to retain residual memory about the data from a distant layer. Inspired by this phenomenon, we conduct an empirical analysis to investigate how the information of trained data in the previous iterations linger over time in the whole training process.

Intuitively, to eliminate the influence of a data to model parameters, we have to first determine what changes it has made. Assume there are $B$ blocks ($\{D_1, D_2, \ldots, D_B\}$) for training, and the intermediate model with $i$-th block is $M_i = F(\{D_1, D_2, \ldots, D_i\})$ where $1 \leq i \leq B$. 
\begin{mydef}\label{def:influence}
	\textbf{(Temporal Influence)} The influence caused by training block $D_i$ can be measured by the difference of the two consecutive models $M_i$ and $M_{i-1}$. It can be formalized as $ Inf(D_i \vert M_{i-1}) = M_i \ominus M_{i-1}$, \ie, the influence by $D_i$ under the condition $M_{i-1}$. 
\end{mydef}

\change{In this definition and hereafter, we reshape a model $M_i$ into a vector $\vec{P_i}$ where the elements of $\vec{P_i}$ are the model parameters of $M_i$ in a topological order. The subtraction of two models ($M_i\ominus M_{i-1}$) is converted into two vectors ($\vec{P_i}-\vec{P_{i-1}}$). The detailed definition is } 
\begin{equation}\label{equ:model-add-sub}
\begin{split}
    M_a\oplus M_b = \vec{P}(M_a)+\vec{P}(M_b);\  M_a\ominus M_b = \vec{P}(M_a)-\vec{P}(M_b)
\end{split}
\end{equation}
\revise{Where $\vec{P}(M_a), \vec{P}(M_b)$ are the parameters of $M_a, M_b$ in a fixed order.}

The influence $Inf$ is correlated to the input block $D_i$ as well as its prerequisite $M_{i-1}$. 
\revise{Supposing the unlearned point $x_d$ locates in block $D_d$.}
$x_d$ makes no influence to its precedent intermediate models $\{M_1, \ldots, M_{d-1}\}$ since it is never seen by them, \ie, $x_d \notin F^{-1}(M_{d-1})$.
But it imposes certain influence to the successor models from $M_{d+1}$ to $M_B$. 
In order to unlearn data $x_d$ from the model, we make the first effort to measure its temporal influence over training. 
Assume that two training processes are performed on the datasets $\{D_1, \ldots, D_{d-1}, D_d, D_{d+1}, \ldots, D_B\}$ and $\{D_1, \ldots, D_{d-1}, D'_d, D_{d+1}, \ldots, D_B\} (D'_d = D_d\backslash \{x_d\})$, we obtain $M$ and $M'$, respectively. According to Definition~\ref{def:naive}, $M'$ is actually the naively unlearned model of $M$ without data $x_d$.
By comparing the models after the $d$-th block in the two training processes, we can measure the difference of models w/ and w/o data $x_d$. 
Therefore, we put forward the concept \emph{temporal residual memory}. 
Different from the residual memory in ResNet that represents the influence over layers, our temporal residual memory is the influence of a data point to the parameters of intermediate models.
After $t$ blocks from block $D_d$, the influence of $x_d$ still partially contributes to $Inf(D_{d+t} | M_{d+t-1})$. So we have the following definition.
\begin{mydef}\label{def:residual}
	\textbf{(Temporal Residual Memory)} With a deep learning model $M$ and its unlearned model $M'$ without data $x_d (x_d \in D_d)$, the temporal residual memory of data $x_d$ after $t$ blocks can be computed as $\Delta_t = \Arrowvert Inf(D_{d+t}|M_{d+t-1}) - Inf(D_{d+t}|M'_{d+t-1}) \Arrowvert_1,~ where~ 0 \leq t \leq B-d$.
\end{mydef}

\change{Here we use $L_1$ distance (Manhattan Distance~\cite{Manhattan-distance}) to quantify the difference ($\Delta_t$) between two $Inf$ vectors. The $L_1$ distance is linear dependent on every pair of elements (\eg, $distance=|w_1-w_2|+|b_1-b_2|$).}
The rationale with this measure is that the output of one layer is also linear dependent on the weights and bias of the layer (\eg, $Y=wX+b$, regardless of activation function)~\cite{u27-influence2017icml}.

As shown in Figure~\ref{fig:residual}, we plot three curves to visualize the changes of $\Delta$ over time in three different settings on the MNIST dataset. \revise{In Figure~\ref{fig:curve100}, $B=600$ and $d=150$ (block $D_d$ contains the unlearned data). In Figure~\ref{fig:curve200}, $B=300$ and $d=50$. In Figure~\ref{fig:curve500}, $B=120$ and $d=30$.}
Obviously, the $\Delta$ value is 0 before block $D_d$.
It is observed that the $\Delta$ curves have a salient trend where they go with a sharp drop in the first several blocks, and then with a mid drop. After a specific number of blocks, the curves stay nearly stable, with small fluctuations around an average value.
The fluctuations are partially due to the randomness (\eg, dropout) of DNNs. 
Because they are independent and identically distributed, we treat them as one type of white noise~\cite{whitenoise}. 
From these curves, we conclude that the unlearned data ceases to influence the parameters in a sufficiently-long time frame along with model training.
Based on these phenomena, we consider whether it is possible to reduce the effort of retraining by reusing part of model parameters, and then implement a model stitching to construct the final unlearned model.

\section{The \approach Approach}\label{sec:approach}

Inspired by the temporal residual memory existing in training, we propose to store intermediate models produced in training, reduce the retraining efforts and reuse unaffected stored models to accomplish an unlearning task.  
Figure~\ref{fig:overview} shows how \approach is integrated into an iterative training process. \revise{Algorithm~\ref{alg:whole} describes the start-to-finish process of \approach.} \change{The four parts correspond to the next four subsections. Stored training in part 1 logs the model's parameters after every \revise{block}. For $B$ blocks in total, the stored training generates and saves $B-1$ intermediate stored models and one finalized model (Original Model). Part 2 quantifies the residual memory of the unlearned data $x_d$. An on-the-fly trend analysis in part 3 determines whether the residual memory can be ignored. Part 4 constructs the unlearned model which unlearns $x_d$ through model stitching. In this section, all variables with ``$'$'' are from the unlearning process.}

\subsection{Stored Training}\label{sec:approach:training}

\revise{\approach divides the entire dataset into disjoint blocks. The dividing process needs to be as uniform as possible, that is, each block has approximately the same number of data points with the same label. Blocks are fixed after the dividing process, so the ``Shuffle'' option should be turned off during training.}

To quantify the influence of one \revise{block} to model parameters, we adopt a stored training to record intermediate models for each block. \revise{Stored training treats every block as a complete dataset for multi-epoch training.}
We assume that there are $B$ blocks in dataset $D$. For each block $D_i \in D$, we save model parameters as $\vec{P}_i$ after training $D_i$. The finalized trained model is $M$, so the parameters of $M$ are actually $\vec{P}_B$.

Different from the work~\cite{u3-DBLP:journals/corr/abs-1912-03817}, a big dataset is divided into several small sets, and the intact model is replaced by many small models trained on different small sets. Although stored training divides the dataset into multiple \revise{blocks}, we still perform the training on the complete dataset $D$ and get an intact model. \change{This makes our approach be more close to the normal training process and achieve higher model accuracy.}

\subsection{Quantification of Residual Memory}

The next challenge is to determine when we should stop retraining, a.k.a., residual memory of unlearned data can be ignored. To tackle it, we introduce parameter vector to represent a stored model during training. 
Given a stored model $M_{k-1}$, we suppose that it contains $m$ parameters including weights, bias, and so on. 
We use $\vec{P}_{k-1}$ to denote the parameter vector, and $\vec{P}_{k-1} \in \mathbb{R}^m$. The vector is a point in $m$-dimensional space. After adding block $D_k$, we train it and update the model to $M_k$ and obtain another parameter vector of $\vec{P}_k$. The directed line segment of two points forms a new vector. We use such a vector in $m$-dimensional space to reflect the change between two consecutive stored models as follows.
\begin{equation}\label{equ:splice-1}
\begin{split}
    \vec{V}_{k} = \vec{P}_{k} - \vec{P}_{k-1}
\end{split}
\end{equation}
\revise{Since we perform a retraining from the block $D_d$ which contains the unlearned data $x_d$}, we can generate new stored models without $x_d$. Assume that the retrained models are $M'_i$ where $d \leq i \leq n$, the parameter vector for each retrained stored model can be represented as $\vec{P'}_i$. 
Therefore, we calculate the update direction of two consecutive parameter vectors as follows.
\begin{equation}\label{equ:splice-2}
\begin{split}
    \vec{V'}_{k} = \vec{P'}_{k} - \vec{P'}_{k-1}
\end{split}
\end{equation}

If the deleted data has little impact on the current data slice training, the directions and sizes of the two vectors are almost the same. Otherwise, the impact of the deleted data can be reflected in the difference of the parameter change vectors. 
We consider that model parameters (either weights or bias) appear in a linear manner during the prediction process, such as $Y=wX+b$ \change{(without considering the activation function)}. So we introduce the L1-norm of vector distance between $\vec{V}_{k}$ and $\vec{V'}_{k}$ to evaluate the difference. 
\begin{equation}\label{equ:splice}
\begin{split}
  \Delta_k = \Vert \vec{V}_{k} - \vec{V'}_{k} \Vert_1
\end{split}
\end{equation}
\change{Here we calculate the distance between two model updates to 
reflect the influence of the previously unlearned data.} When this value $\Delta_k$ becomes very small, it means that the two update vectors almost overlap, and the influence of the deleted data in $D_d$ can be ignored for model parameters. 

\subsection{On-the-fly Trend Analysis}\label{sec:approach:dfa}

It is challenging to determine when $\Delta$ is small enough \change{for stop retraining} due to its non-determinism with unpredictable fluctuations. 
As described in the empirical study in Section~\ref{sec:problem}, we plot three figures (Figure~\ref{fig:residual}) to present the variation tendency of $\Delta$. 
The value of $\Delta$ is very high near the unlearned \revise{block position $d$}, which means the deleted data has a great influence on the model. Then it drops rapidly, and finally oscillates around a value. This variation curve is roughly consistent with the shape of power-law decaying functions, \ie, $Y=a\cdot X^{-h}+b$. 
By fitting the sequence of $\Delta$ into a power-law decaying function, we can determine whether the sequence stays stable and the value of $\Delta$ will not change considerably. 

To eliminate the noise in the curve of $\Delta$, we introduce \emph{detrended fluctuation analysis (DFA)}~\cite{dfa}. DFA is used to determine the statistical self-affinity of a signal. \change{If a time series has a gradual structure of non-randomly decreasing autocorrelation, DFA can quantitatively analyze the slowness of the decay of these correlations and reflect it in the form of DFA index.} DFA is efficient for analyzing time series that appear to be long-memory processes such as power-law decaying autocorrelation function. Here, the sequence $\{\Delta_d,\cdots,\Delta_{d+t}\}$ can be seen as the time series data, \change{where $t$ is the number of blocks to be retrained, starting from block $D_d$. The DFA function computes an exponent $h$ in the power-law decay function for time series data.}

After obtaining the exponent using DFA method, we need to solve the parameters $a$ and $b$ to fit the sequence of $\Delta$ to a power-law decaying function. For this purpose, we adopt the least square method and perform the following optimization.
\begin{equation}\label{equ:splice}
\begin{split}
    \arg \min_{a,b} ~(a\cdot x^{-h} + b) - \Delta_x
\end{split}
\end{equation}
Where $x$ is the \revise{order of block}. When the sequence reaches a stationary state, it should have a very small change trend at the right end $d+t$, that is, the absolute value of the derivative of the power-law function $Y$ should be very small. Therefore, we use the absolute value of this derivative as a measure to determine stationarity of the $\Delta$ sequence, \ie, $\frac{\partial (a\cdot x^{-h} + b)}{\partial x}~=~a\cdot (-h) \cdot x^{-h-1}$.
If the derivative is less than a certain value $\varepsilon$, the sequence reaches the stationary state and we terminate the retraining. 
Otherwise, we let $t = t + 1$ and continue the retraining.

\begin{algorithm}[t]
	\caption{Algorithm of \approach}\label{alg:whole}
	\KwIn{$\vec{P}_k$: original model parameters after training the $k$-th block; $B$: total number of blocks; $d$: \revise{the number of block which has the deleted data}; $\varepsilon$: a constant to determine the stationarity. }
	\KwOut{$M'$: unlearned model with updated block $D_d$.}
	$M' \leftarrow \vec{P}_{d-1}$;  \\
	$\vec{P'}_{d-1}\leftarrow \vec{P}_{d-1}$;  \\
	Update block $D_d$; \quad$\triangleright$ delete data from block $D_d$ \\
	\For{$t\leftarrow 0$ \textbf{to} $B-d$}{  
		$M' \leftarrow $ train $D_{d+t}$ on $M'$; \quad$\triangleright$ t is retrain length  \\
		$\vec{P'}_{d+t} \leftarrow M'$;   \quad$\triangleright$ get parameters from model $M'$\\
		$\vec{V}_{d+t} = \vec{P}_{d+t} - \vec{P}_{d+t-1}$; \quad$\triangleright$ temporal influence\\
		$\vec{V'}_{d+t} = \vec{P'}_{d+t} - \vec{P'}_{d+t-1}$; \\
		$\Delta_{d+t} = \Vert \vec{V}_{d+t} - \vec{V'}_{d+t} \Vert_1$; \quad$\triangleright$ temporal residual memory\\
		$h = $ DFA( $\{ \Delta_{d}, \Delta_{d+1}, \ldots, \Delta_{d+t}\}$ ); \\
		$Y(x) = a\cdot x^{-h}+b$; \\
		$f(x) = \partial Y(x)/\partial x = a\cdot (-h)\cdot x^{-h-1}$; \\
		$a,b = \arg \min( Y(x)- \Delta_x )$; \quad$\triangleright$ use Least-square\\
		$p = f(d+t)$; \\
		\If{$|p|<\varepsilon$}{\textbf{break};} 
	}
	\revise{$M' \leftarrow M' \oplus (M_{B}\ominus M_{d+t})$;} \quad$\triangleright$ model stitching\\
	\revise{Update [$\vec{P}_{d}, \cdots, \vec{P}_{d+t}$] with [$\vec{P'}_{d}, \cdots, \vec{P'}_{d+t}$];} \\
	\Return{$M'$}
\end{algorithm}

\subsection{Model Stitching}\label{sec:approach:stitch}

Finally we need to use stored models to stitch together the impact of the last part of blocks.
\revise{For example, a trained model ($M_B$) can also be represent jointly by an intermediate model ($M_d$) and model stitching, that is}
\begin{equation}\label{equ:stitch-example}
\begin{split}
\revise{M_B = M_d \oplus (M_B \ominus M_d)}
\end{split}
\end{equation}
We suppose the deleted data belongs to block $D_d$, and we update $D_d$ to $D_d'$ according to the unlearn request. The unlearned model needs to retrain on the new dataset $\{D_1, \cdots, D_{d-1}, D'_d, D_{d+1}, \cdots, D_B\}$. We first construct the initial unlearned model $M'$ which has the same parameters as $\vec{P}_{d-1}$. Assuming the retrained interval is $t$, we next train $M'$ on blocks $\{D_d', D_{d+1}, \cdots, D_{d+t}\}$ and update $M'$. Finally, we stitch the influence of \revise{blocks} $\{D_{d+t+1}, \cdots, D_B\}$ on model parameters to the model $M'$, that is 
\begin{equation}\label{equ:splice}
\vspace{-2mm}
\begin{split}
M'\leftarrow M'\oplus (M_{B}\ominus M_{d+t}) 
\end{split}
\vspace{-2mm}
\end{equation}

\revise{As shown in Equation~\ref{equ:model-add-sub}, $\oplus$ and $\ominus$ represent the addition and subtraction between two models, and we treat them as two matrices. The result is a new matrix, and also a new model.} 

\subsection{Algorithm Summary and Analysis}\label{sec:approach:summary}

Algorithm~\ref{alg:whole} presents the workflow of \approach. 
We initialize the unlearned model $M'$ from line 1 to 3, and update the block $D_d$. Line 4 to 16 is to retrain model and determine stationarity on-the-fly. \change{We enumerate the retrained block length $t$ at line 4.} At line 5, we train model $M'$ on block $b_{d+t}$. The parameters of $M'$ are updated and accessed by $\vec{P}_{d+t}'$ at line 6.
Then we calculate the L1-norm distance between two model updates (line 7-9) to measure the residual memory of unlearned data. 
Based on the DFA method, we solve the exponent $h$ at line 10. 
As the curve of $\Delta$ conforms to a power-law decaying function, we construct it at line 11 and compute its derived function $f(x)$ at line 12. 
We adopt the least square method to solve values $a$ and $b$ to fit the function at line 13. 
The derivative is computed at line 14 at the right end $d+t$, and if its absolute value is smaller than $\varepsilon$, we stop the retraining. 
The model is then stitched with the remaining stored models, and last we obtain the unlearned model in Line 17. Line 18 updates the array $\vec{P}$ for subsequent deletions. 

\change{The training part at line 5 needs to be repeated for $t$ times. The maximum $t$ is $B-d$, however, we have an early termination condition at line 15, 16. According to the experiment results, $t=0.01\sim0.08\times B$ when unlearning one data point. The training cost of our approach is about 1\%$\sim$8\% of the naive method. The non-training costs (Line 6$\sim$16) mainly stem from parameter-scale arithmetic calculations and DFA with a small ($<B$) sequence, which are negligible compared to the retrain time.}

	\section{Evaluation}\label{sec:eval}

\begin{figure}[t]
	\centering
	\includegraphics[width=0.33\textwidth]{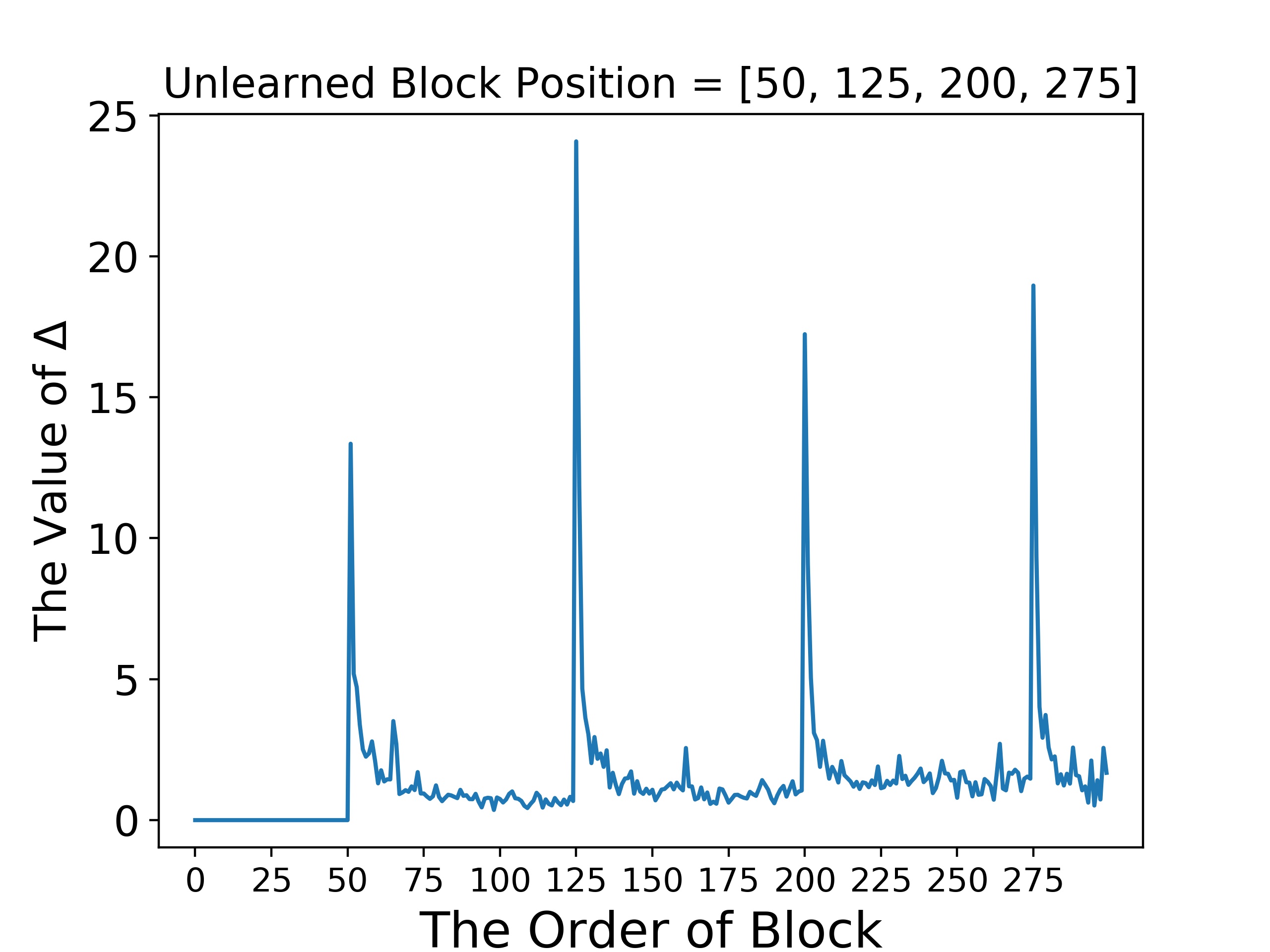}
	\caption{The variation trend of residual memory $\Delta$ when deleting multiple blocks in one shot on MNIST ($B=300$).} \label{fig:mul-del-200}
\end{figure}
\begin{figure}[t]
	\centering
	\includegraphics[width=0.33\textwidth]{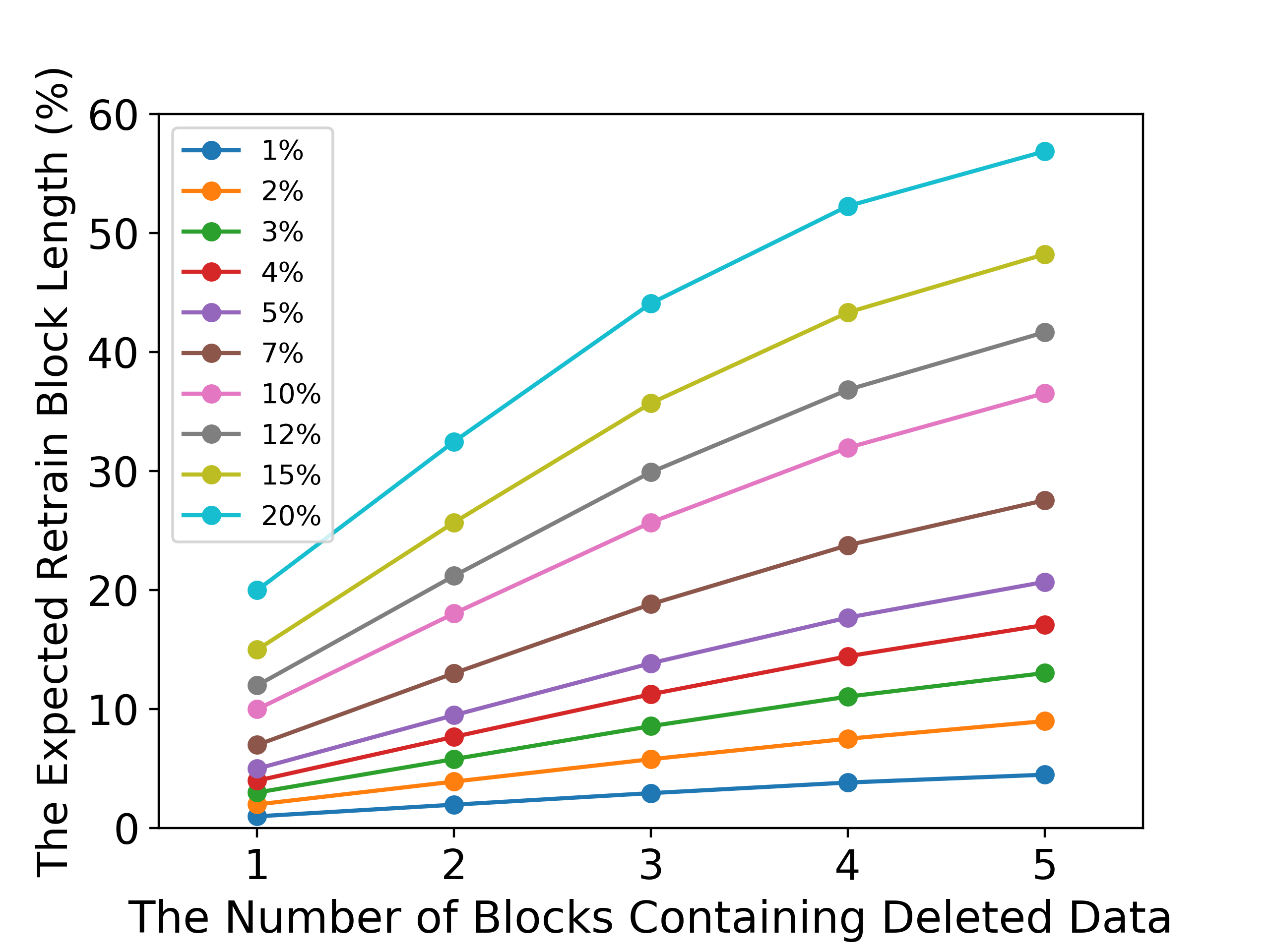}
	\caption{The expected percent of retained blocks for multi-blocks deletion. X-axis is the number of deleted blocks. The number in legend is the retrain length for deleting one block.}
	\label{fig:mul-del-expectation}
\end{figure}

\begin{table*}[t]   
	\caption{Evaluations of the MNIST dataset on LeNet-5. We evaluate our method under different \revise{unlearned data numbers} and deletion positions. The test accuracy of original trained model reaches 99.10\%. \revise{Here $B=500$.} \change{``No. of Un- Data'' is the number of unlearned data. ``Unlearned Position'' means which block has unlearned data in a deletion request, such as 1st block.} ``$\varepsilon$'' provides the condition for stopping retraining, and we enumerate $\varepsilon = 0.1 \sim 0.04$ here. ``Acc.'' means the accuracy of unlearned model on test dataset. ``Con.'' means the consistency between our unlearned model and the naive unlearned model on test dataset. \revise{``Speed-up($\times$)'' is the retraining time speed-up compared to naive unlearning (1$\times$).} \revise{For ``Naive'' unlearning, the speed-up is 1$\times$, and ``Con.'' is 100\%, which also apply to Table~\ref{tbl:svhn-vgg11}, \ref{tbl:cifar10-resnet18}, \ref{tbl:purchase-dnn}, \ref{tbl:imagenet-resnet50}.}}\label{tbl:mnist-lenet5} 
	\centering
	\scriptsize
	\resizebox{1\textwidth}{!}{
		\begin{tabular}{ccccccccccccccc} \toprule
			No. of & Unlearned & \multicolumn{3}{c}{\approach $\varepsilon = 0.1$} & \multicolumn{3}{c}{\approach $\varepsilon = 0.08$} & \multicolumn{3}{c}{\approach $\varepsilon = 0.06$} & \multicolumn{3}{c}{\approach $\varepsilon = 0.04$} & Naive \\ \cline{3-15}
			Un- Data & Position & Acc.(\%) & Con.(\%) & \revise{Speed-up($\times$)} & Acc.(\%) & Con.(\%) & Speed-up($\times$) & Acc.(\%) & Con.(\%) & Speed-up($\times$) & Acc.(\%) & Con.(\%) & Speed-up($\times$) & Acc.(\%) \\ \toprule
			\multirow{3}{*}{1} 
			&   1st & 98.95 & 99.85 & 66.67 & 98.99 & 99.90 & 58.82 & 99.02 & 99.92 & 52.63 & 99.03 & 99.95 & 45.45 & 99.08\\ 
			& 300th & 98.94 & 99.86 & 66.67 & 98.97 & 99.91 & 55.55 & 99.00 & 99.90 & 52.63 & 99.03 & 99.93 & 45.45 & 99.06\\
			& 600th & 98.93 & 99.84 & 66.67 & 98.97 & 99.88 & 58.82 & 99.01 & 99.92 & 52.63 & 99.02 & 99.95 & 47.62 & 99.07\\ \midrule 
			\multirow{3}{*}{60} 
			&   1st & 97.72 & 99.28 & 30.30 & 97.85 & 99.42 & 27.03 & 97.93 & 99.48 & 22.73 & 98.06 & 99.60 & 18.52 & 98.43\\ 
			& 300th & 97.66 & 99.24 & 35.71 & 97.80 & 99.38 & 30.30 & 97.90 & 99.48 & 23.26 & 98.02 & 99.52 & 19.23 & 98.41\\  
			& 600th & 97.68 & 99.27 & 33.33 & 97.82 & 99.41 & 28.57 & 97.91 & 99.49 & 24.39 & 97.95 & 99.54 & 20.00 & 98.40\\ \bottomrule
			\end{tabular}
	}
\end{table*}

To evaluate the efficacy of \approach, we conduct comprehensive and extensive experiments on five mainstream datasets with five canonical deep learning models. Through the experiments, we intend to answer the following questions.

\begin{enumerate}[leftmargin=*,label=\textbf{RQ$\arabic*$.}]
	\item How effective is \approach in deep unlearning from three aspects of consistency, accuracy and speed-up?
	\item How effectively can \approach pass the verification of deep unlearning?
	\item How is the comparison between \approach and the state-of-the-art unlearning approach?
\end{enumerate}

\subsection{Experimental Setup}\label{sec:eval:setup}

\noindent\textbf{Datasets}. We evaluate our method on five following datasets.
\begin{itemize}[leftmargin=*]
	\item \noindent\textbf{MNIST}~\cite{mnist} is an image dataset of handwritten digits, containing 60,000 training data and 10,000 test data. It has 10 classes of digit 0 to 9. Each image is 28$\times$28 grey-scale.
	\item \noindent\textbf{CIFAR-10}~\cite{cifar10} contains 50,000 training data and 10,000 test data. It has 10 classes of vehicles or animals, such as plane, car, bird, cat, dog, and so on. Each image is a 32$\times$32 RGB. 
	\item \noindent\textbf{SVHN}~\cite{u18-SVHN} is a real-world street view house numbers dataset collected by the Google Maps service. It has 604,388 training data and 26,032 test data of 10 classes. Each image is 32$\times$32 RGB.
	\item \noindent\textbf{Purchase}~\cite{u19-purchase} is an online shoppers' purchasing intention dataset. It has 12,330 instances and 11,097 training samples. Each sample is multivariate with 18 attributes.
	\item \noindent\textbf{ImageNet}~\cite{imagenet} is a large-scale visualization database for visual object recognition. We use ILSVRC 2012, one of its subset. It contains 1,281,167 training data, 50,000 validation data, and 100,000 test data of 1,000 classes. Each image is 224$\times$224 RGB.
\end{itemize}

\noindent\textbf{The Choice of Models and Datasets.} 
To train qualified models for unlearning, we choose different model architectures which are already proved to be effective in a certain dataset. 
In particular, we choose the LeNet-5 model to train it on MNIST, the ResNet-18 model on CIFAR-10, the VGG-11 model for SVHN, a DNN model with one hidden layer for Purchase and the ResNet-50 model on ImageNet. \change{We evaluate MNIST and CIFAR-10 because they are two basic commonly used datasets. We test SVHN and Purchase because they contain users' house addresses and shopping records, which are closely related to their privacy. ImageNet is chosen to test the adaptability on a large-scale and more complex dataset.}

\noindent\textbf{Baseline.}
\change{Naive unlearning serves as the baseline method in the experiments. As aforementioned, it is perfect in unlearning but catastrophic in time efficiency. Our goal is to approximate its unlearning performance with the lowest possible overhead.}

\noindent\textbf{Implementation.}
We implement our unlearning approach on top of \textsc{Pytorch}~\cite{pytorch}. We select the Adam optimizer~\cite{adam}. 
All our experiments are conducted on a Linux server with 16 Intel(R) Xeon(R) CPUs of E5-2620 and 32GB memory, 2 NVIDIA GM200 [GeForce GTX TITAN X] GPUs and 1 ASPEED Video AST2400 GPU.

\subsection{Effectiveness}\label{sec:eval:effective}

In this section, we evaluate \approach with three criteria--accuracy, consistency, and \revise{speed-up} under different $\varepsilon$ values, unlearned data numbers, and unlearned positions. Naive unlearning serves as a baseline method, and the consistency and speed-up are computed based on the baseline. \revise{Without loss of generality, our experiments assume that when deleting several data points, they are from the same block. For deletions from multi-blocks, it can be accomplished with multiple one-block deletions.}

\revise{Figure~\ref{fig:mul-del-200} shows the change curve of $\Delta$ in the situation of deleting data points from multi-blocks. When deleting each block, the change curve of $\Delta$ (such as 50th$\sim$124th-block) still follows the rule as shown in Figure~\ref{fig:residual}. So the problem of deleting data from multi-blocks can be reduced to one-block deletions. }
\revise{Figure~\ref{fig:mul-del-expectation} lists the expected retrain length (expressed as $\%$ of the total number of blocks) under multi-blocks deletions, which are theoretical calculations. The number in legend ($1\%\sim 20\%$) represents the retrain cost of deleting data from one block. The point (x=5, y=56.9\%) in the $20\%$-line means if one-block deletion needs $20\%\times B$ retrain cost, unlearning data from 5 blocks expects $56.9\%\times B$ cost. For simplicity and to discover more essential findings, our evaluations only focus on unlearning data from one block.}

\subsubsection{Unlearning LeNet-5 on MNIST}
Table~\ref{tbl:mnist-lenet5} presents the evaluations on MNIST using LeNet-5. \revise{We set $B=500$.} 
We choose different unlearned positions to study whether our method is affected by deletion positions. It is also evaluated under different $\varepsilon$ values, corresponding to different termination conditions.

\revise{When unlearning 1 data point, \approach can reach 99.03\% accuracy, very close to 99.07\% in naive unlearning, and achieve 45.45$\times$ speed-up.}
From deleting 1 data point to 60, the accuracy of unlearned model reduces 1.25\% under $\varepsilon=0.1$, 1.01\% under $\varepsilon=0.04$, and 0.66\% under naive unlearning on average. 
\revise{Next we unlearn data from 1st, 300th, 600th-block. Results show that there are no obvious gaps in accuracy, consistency and speed-up when deleting data from different block. It indicates that \approach is not limited to unlearning data in specific blocks.} 
A tighter termination condition (a smaller $\varepsilon$) needs more retraining, and increases the accuracy and consistency. When deleting 60 data, \approach reaches 97.69\% accuracy, 99.26\% consistency with 3.03\% cost (equivalent to 33.0$\times$ speed-up) on average under $\varepsilon=0.1$, and 98.01\% accuracy, 99.55\% consistency with 5.2\% cost (19.2$\times$ speed-up) under $\varepsilon=0.04$. 

\begin{remark}
\approach reaches close accuracy to naive unlearning with very little overhead. Different unlearned positions have little effect on model performance. \approach is not limited by the location of deleted data. \change{Additionally, $\varepsilon$ can be used to tune the performance of unlearning. A smaller $\varepsilon$ leads to a higher consistent and accurate unlearning but causes more retraining cost.}
\end{remark} 

\begin{figure*}[t]
	\begin{subfigure}{.32\textwidth}
		\centering
		\includegraphics[width=\linewidth]{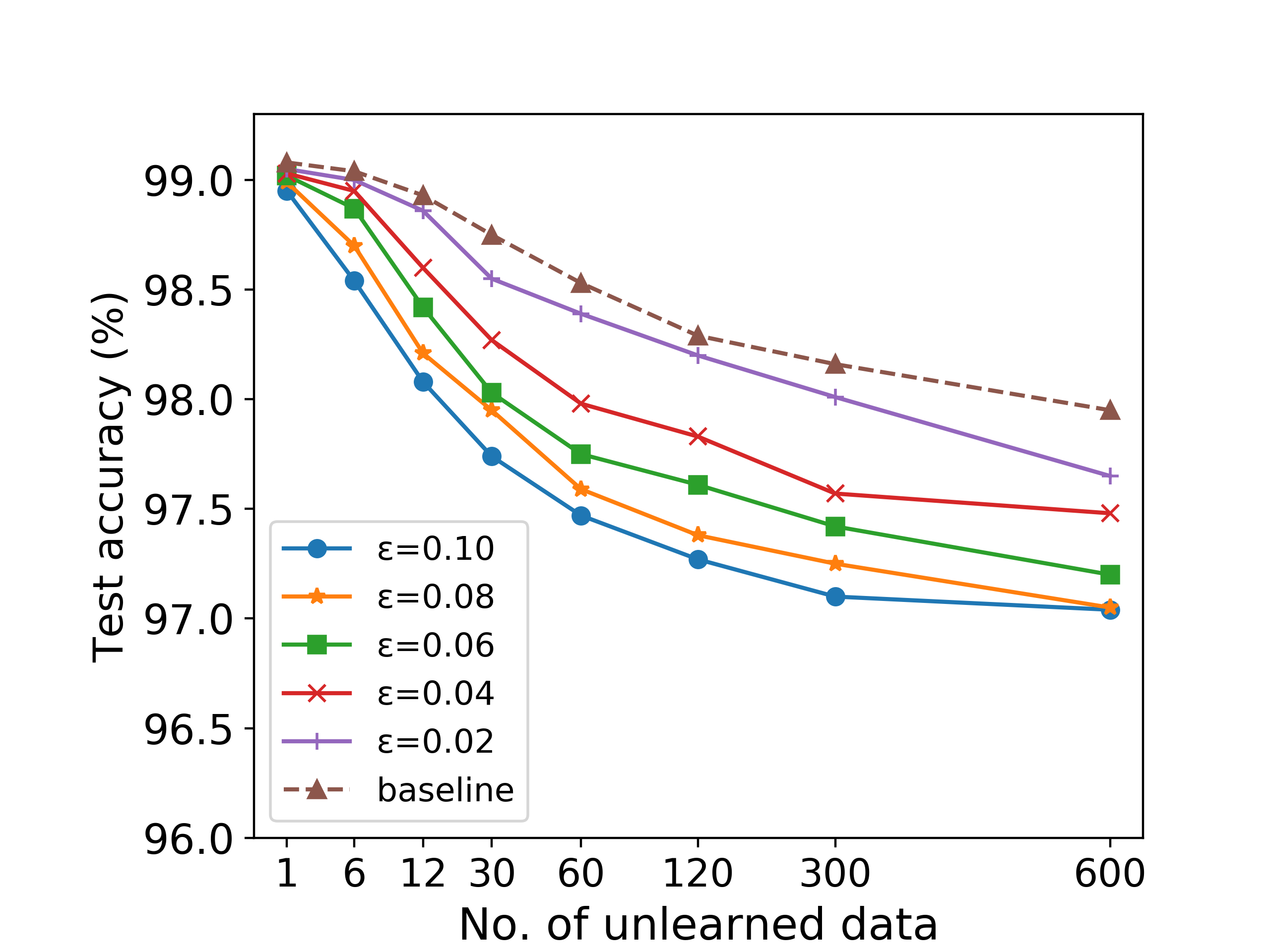}  
		\caption{The test accuracy of \approach under different $\varepsilon$ and the baseline unlearning.}\label{fig:mini-del-acc-100}
	\end{subfigure}
	\hfill
	\begin{subfigure}{.32\textwidth}
		\centering
		\includegraphics[width=\linewidth]{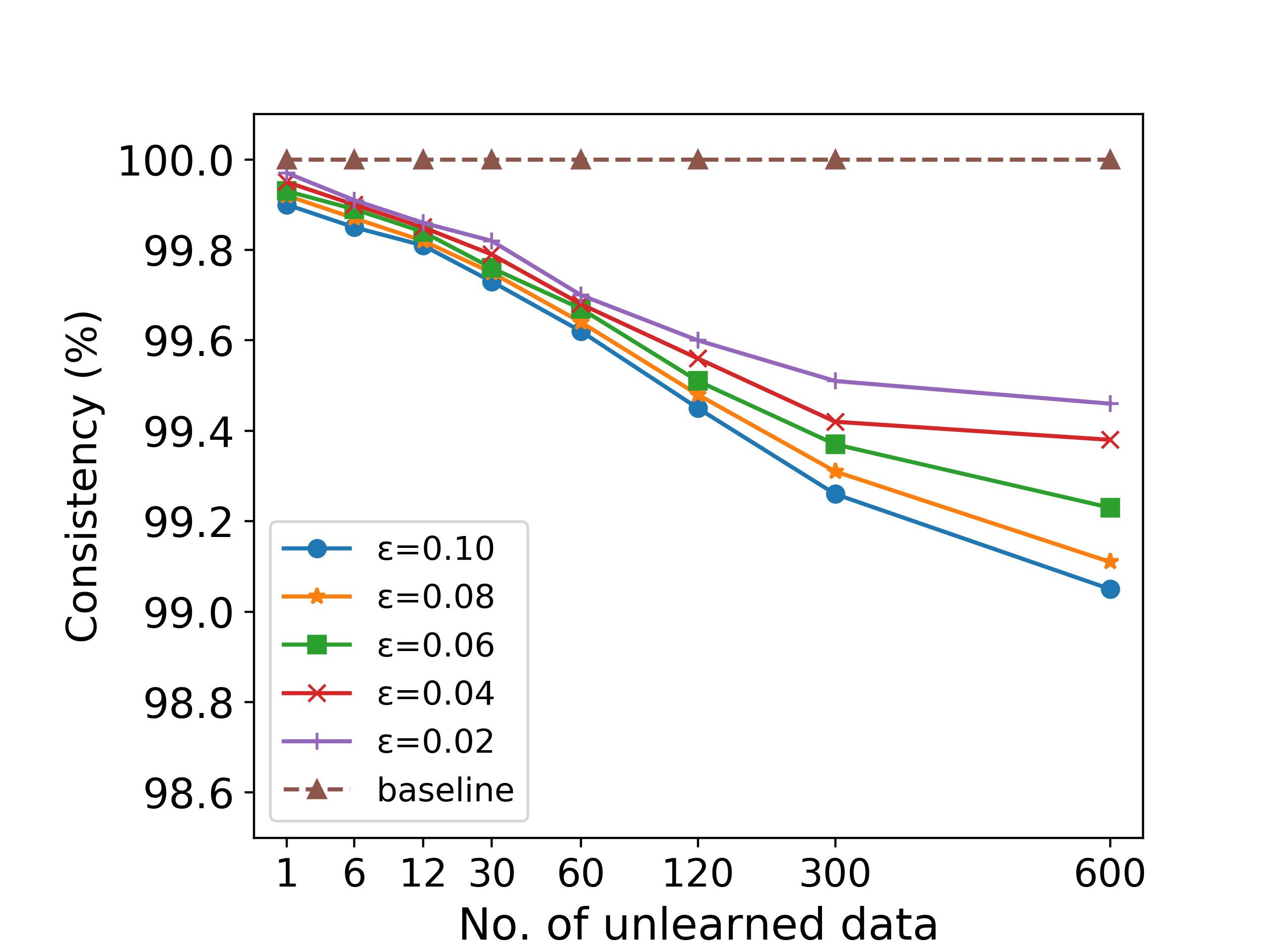}  
		\caption{The consistency of \approach with the baseline model under different $\varepsilon$.}\label{fig:mini-del-con-100}
	\end{subfigure}
	\hfill
	\begin{subfigure}{.32\textwidth}
		\centering
		\includegraphics[width=\linewidth]{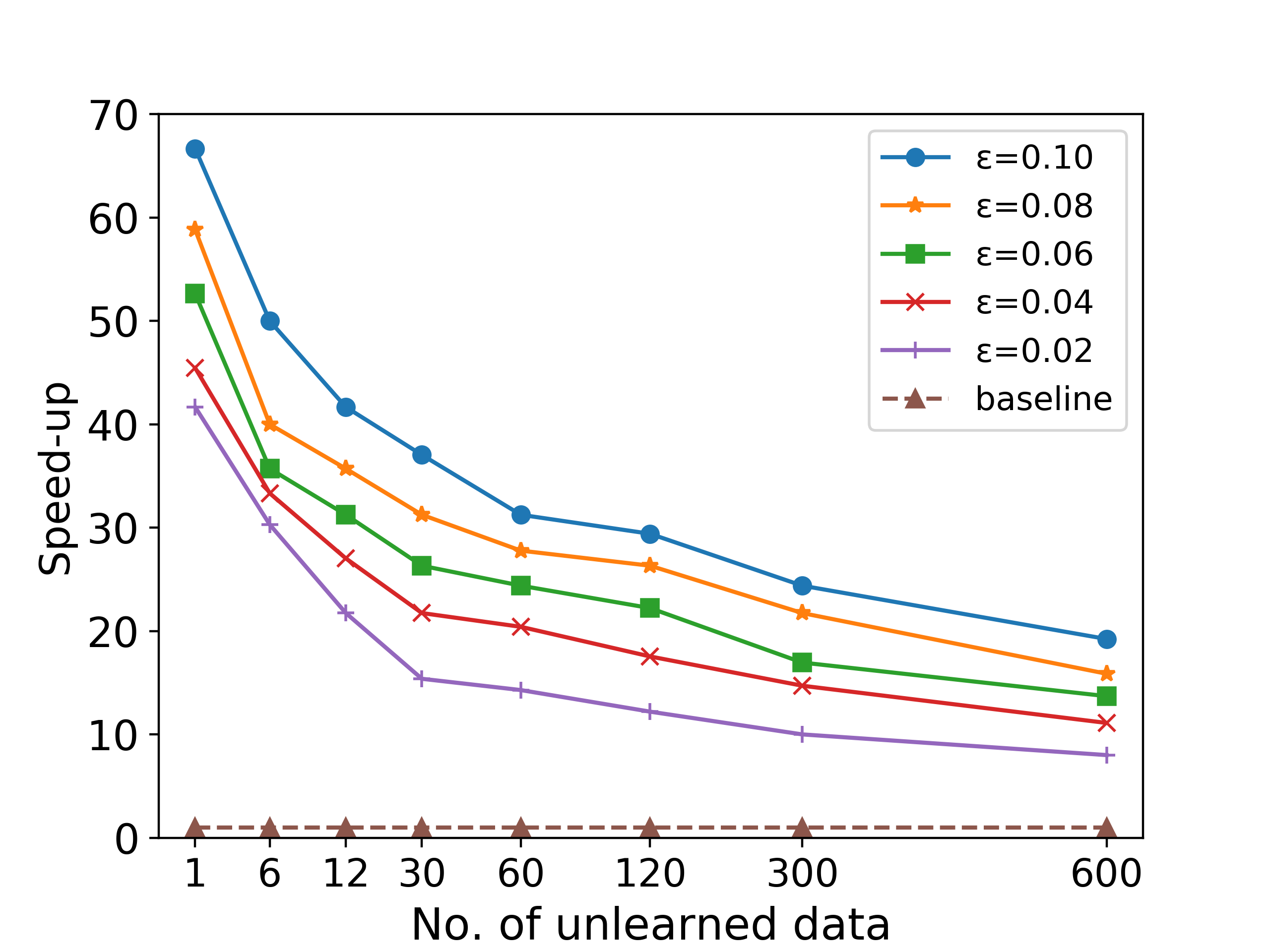}  
		\caption{\revise{The speed-up of \approach is compared to the naive unlearning (1$\times$).}}\label{fig:mini-del-cost-100}
	\end{subfigure}
	
	\caption{\change{Evaluations of unlearning a small amount of data on MNIST ($B=1,000$). The unlearned data starts from the 10\% (100th) position consecutively. The x-axis is the number of unlearned data points, varying from 1 (0.0017\%) to 600 (1\%).}}
	\label{fig:mini-del}
\end{figure*}

\revise{Next we study the impact of unlearning different data numbers. Figure~\ref{fig:mini-del} explores the situations from unlearning only 1 data point (0.00167\%) to 600 points (1\%).
Based on different $\varepsilon$ from 0.1 to 0.02, the accuracy, consistency and speed-up curves all go down as unlearning more data. For the baseline, the consistency and speed-up are compared to itself, so these two curves are always 100\% and 1$\times$. }

\change{
Specifically, the accuracy of \approach is very close to the baseline as Figure~\ref{fig:mini-del-acc-100}. The maximum accuracy gap is 1.06\% at $\varepsilon=0.10$, and only 0.30\% at $\varepsilon=0.02$. 
In Figure~\ref{fig:mini-del-con-100}, our method all reaches over 99\% consistency. 
In Figure~\ref{fig:mini-del-cost-100}, we get 8$\sim$19.2$\times$ speed-up at 600 unlearned data for $\varepsilon=0.02\sim0.1$, and 14.3$\sim$31.2$\times$ speed-up at 60 unlearned data, and 30.3$\sim$50$\times$ speed-up at 6 unlearned data.
When unlearning only 1 data point, our method reaches 99.03\% accuracy (0.07\% lower than the original model) and 45.5$\times$ speed-up. }

\begin{remark}
Accuracy and consistency have a negative correlation to unlearned data number and $\varepsilon$ value, while speed-up increases with a larger $\varepsilon$. More unlearned data incurs more degradation to the model (\ie, accuracy) and leaves more residual memory that is not eliminated (\ie, consistency) whilst requiring more retraining. 
\end{remark}

\begin{table*}[t] 
\caption{Evaluations of the SVHN dataset on VGG-11. The test accuracy of the original model reaches 95.40\%. \revise{We set $B=1,000$}. ``\% of Un- Data'' refers to the percents of unlearned data in the training dataset. ``Unlearned Position'' means which block contains the unlearned data. ``$\varepsilon$'' specifies the condition for stopping retraining, and we enumerate $\varepsilon = 0.1 \sim 0.05$ here.}
	\label{tbl:svhn-vgg11}
\centering
\scriptsize
        \begin{tabular}{cccccccccccc} \toprule
              \% of & Unlearned& \multicolumn{3}{c}{\approach $\varepsilon = 0.1$ } & \multicolumn{3}{c}{\approach $\varepsilon = 0.08$} & \multicolumn{3}{c}{\approach $\varepsilon = 0.05$} & Naive \\ \cline{3-12}
              Un- Data & Position & Acc.(\%) & Con.(\%) & Speed-up($\times$) & Acc.(\%) & Con.(\%) & Speed-up($\times$) & Acc.(\%) & Con.(\%) & Speed-up($\times$) & Acc.(\%) \\ \toprule
              \multirow{3}{*}{0.01\%} &
                 1st & 94.88 & 98.66 &  9.26 & 94.95 & 98.90 & 8.13 & 95.10 & 99.04 & 6.17 & 95.22  \\  
              & 10th & 94.80 & 98.78 &  9.71 & 94.87 & 99.01 & 8.13 & 94.93 & 99.07 & 6.33 & 95.15  \\
              &100th & 94.73 & 98.90 & 10.53 & 94.90 & 99.05 & 8.47 & 94.92 & 99.13 & 6.71 & 95.13  \\ \midrule 
              \multirow{3}{*}{0.1\%} &
                 1st & 94.63 & 97.97 & 6.33 & 94.63 & 98.01 & 5.29 & 94.73 & 98.18 & 4.63 & 94.88  \\ 
              & 10th & 94.54 & 98.02 & 6.58 & 94.76 & 98.26 & 5.38 & 94.79 & 98.35 & 4.93 & 94.91  \\ 
              &100th & 94.34 & 98.55 & 6.80 & 94.56 & 98.60 & 5.56 & 94.68 & 98.83 & 5.10 & 94.82  \\ \bottomrule
        \end{tabular}
\end{table*}

\begin{table*}[t]  
\caption{Evaluations of the CIFAR-10 dataset on ResNet-18. The prediction accuracy of the original trained model is \revise{92.15\%}. \change{We set $B=100$. ``No. of Un- Data'' is the number of unlearned data, from 1 data to 500 points.} ``Unlearned Position'' means which block has unlearned data. ``$\varepsilon$'' provides the condition for stopping retraining. We enumerate $\varepsilon = 0.1 \sim 0.05$.} 
\label{tbl:cifar10-resnet18}
\centering
\scriptsize
        \begin{tabular}{cccccccccccc} \toprule
               No. of & Unlearned & \multicolumn{3}{c}{\approach $\varepsilon = 0.1$} & \multicolumn{3}{c}{\approach $\varepsilon = 0.08$ } & \multicolumn{3}{c}{\approach $\varepsilon = 0.05$} & Naive\\ \cline{3-12}
               Un- Data & Position & Acc.(\%) & Con.(\%) & Speed-up($\times$) & Acc.(\%) & Con.(\%) & Speed-up($\times$) & Acc.(\%) & Con.(\%) & Speed-up($\times$) & Acc.(\%) \\ \toprule
             \multirow{2}{*}{1 } &
               1st & 91.89 & 96.22 &  33.33 & 91.92 & 97.84 &  20.00 & 92.04 & 99.05 &  14.28 & 92.07   \\
             &50th & 91.87 & 96.26 &  33.33 & 91.89 & 97.80 &  20.00 & 92.02 & 99.07 &  14.28 & 92.06   \\ \midrule 
             
             \multirow{2}{*}{20} &
               1st & 91.57 & 93.59 &  11.11 & 91.66 & 95.22 &  7.69 & 91.94 & 97.82 &  5.88 & 92.03   \\    
             &50th & 91.54 & 93.68 &  12.50 & 91.67 & 95.17 &  7.14 & 91.91 & 97.73 &  6.25 & 91.99   \\ \midrule 
            
             \multirow{2}{*}{500} &
               1st & 90.78 & 91.95 &  6.67 & 91.38 & 93.85 &  4.76 & 91.80 & 96.48 &  4.00 & 91.91   \\ 
             &50th & 90.72 & 92.04 &  7.14 & 91.35 & 93.92 &  5.00 & 91.71 & 96.50 &  4.00 & 91.88   \\ \bottomrule
        \end{tabular}
\end{table*}

\subsubsection{Unlearning VGG-11 on SVHN}
Table~\ref{tbl:svhn-vgg11} shows the results on SVHN using a VGG-11 model ($B=1,000$).  
Under 0.01\% and 0.1\% unlearned data, we choose three different unlearned positions. 
\change{
Results show ``1st'' needs a bit more retraining and reaches a bit higher accuracy. Data at the 1st block may have longer influence, which likely affects results. 
Unlearning data at the latter position gets higher consistency. In extreme cases, when deleting the last data, our method and naive method get the same results.
Deleting more data in one shot affects performance. Under $\varepsilon=0.08$, our method reaches 94.91\% accuracy, 98.99\% consistency, and 8.2$\times$ speed-up on average with 0.01\% unlearned, and 94.65\% accuracy, 98.29\% consistency, and 5.4$\times$ speed-up on average with 0.1\% unlearned. 
With the decrease of $\varepsilon$, the accuracy, consistency increase and speed-up decreases. When unlearning 0.1\% data, we get 94.50\% accuracy, 98.18\% consistency with 6.5$\times$ speed-up on average at $\varepsilon=0.1$, and 94.73\% accuracy (close to 94.88\% in naive), 98.45\% consistency with 4.9$\times$ speed-up at $\varepsilon=0.05$.
} 
\begin{remark}
	Unlearning more data needs a bit more cost, but \approach still remains high accuracy and consistency. The accuracy, consistency, cost all go up as $\varepsilon$ decreases. 
\end{remark}

\begin{table}[t]  
\caption{Evaluations of the Purchase dataset on DNN. The prediction accuracy of the original trained model is 97.32\%. We set $B=200$ here. Unlearned position is set to 1st-block.} 
\label{tbl:purchase-dnn}
\centering
\scriptsize
        \begin{tabular}{ccccc} \toprule
             \% of & \multicolumn{3}{c}{\approach $\varepsilon = 0.04$} & Naive   \\ \cline{2-5}
             Un- Data & Acc.(\%) & Con.(\%) & Speed-up($\times$) & Acc.(\%)  \\ \toprule
             \revise{0.01\%} & 97.15 & 99.70 & 33.33 & 97.21 \\ \midrule
             0.1\% & 96.78 & 99.62 & 29.41 & 96.95  \\ \midrule 
               1\% & 95.70 & 99.11 & 17.54 & 96.54  \\ \midrule 
               5\% & 93.67 & 98.79 & 14.49 & 94.81  \\ \midrule 
              10\% & 93.11 & 98.75 & 13.33 & 94.12  \\ \midrule 
        \end{tabular}
\end{table}

\subsubsection{Unlearning ResNet-18 on CIFAR-10}
Table~\ref{tbl:cifar10-resnet18} shows the evaluations on CIFAR-10 using a ResNet-18. 
We specify two unlearned data positions, 1st and 50th-block, and set $B=100$.
We evaluate three $\varepsilon$ values, 0.1, 0.08, 0.05, and three unlearned data numbers, 1, 20, 500. 
\change{
When unlearning only 1 data point, \approach reaches 92.03\% accuracy, close to the original model of 92.15\%, with 14.3$\times$ speed-up at $\varepsilon=0.05$, and 91.88\% accuracy, with 33.3$\times$ speed-up at $\varepsilon=0.1$.
From the perspective of unlearned data number, the accuracy, consistency, and speed-up have a downtrend as unlearning more data. But this hardly affects the performance of \approach. Under $\varepsilon=0.05$, we get 91.92\% accuracy (close to 92.01\% in naive), 97.77\% consistency, 6.1$\times$ speed-up when unlearning 20 data, and 91.75\% accuracy (close to 91.89\% in naive), 96.49\% consistency, 4$\times$ speed-up when unlearning 500 data.
Unlearning different positions has little impact on performance. Deleting data at the 1st position may require a little more retraining and also get a slightly higher accuracy. 
As selecting a smaller $\varepsilon$, the accuracy and consistency also rise and the speed-up drops. 
}

\begin{remark}
	\approach applies to different unlearned positions and data numbers on CIFAR-10. When unlearning only 1 data point, \approach can approximate the original model with a small cost. Different unlearned positions have little effect on the results.
\end{remark}

\begin{table*}[t] 
\caption{Evaluations of the ImageNet dataset on ResNet-50. \revise{$B=1,000$.} The top-1 accuracy is 75.10\% and the top-5 accuracy is 92.48\% on the original model.
``No. of Un- Data'' is the number of unlearned data. ``Unlearned Position'' means which block has the unlearned data. \change{``Top-1.'' is the top-1 accuracy and ``Top-5.'' is the top-5 accuracy on the test dataset.} We select $\varepsilon = 0.1, 0.05$.} 
\label{tbl:imagenet-resnet50}
\centering
\scriptsize
       \begin{tabular}{cccccccccccc} \toprule
               No. of & Unlearned & \multicolumn{4}{c}{\approach $\varepsilon = 0.1$} & \multicolumn{4}{c}{\approach $\varepsilon = 0.05$} & \multicolumn{2}{c}{Naive Unlearned Model}\\ \cline{3-12}
               Un- Data & Position & Top-1.(\%) & Top-5.(\%) & Con.(\%) & Speed-up($\times$) & Top-1.(\%) & Top-5.(\%) &  Con.(\%) & Speed-up($\times$) & Top-1.(\%) & Top-5.(\%)  \\ \toprule
              \multirow{2}{*}{1} &
                  1st & 74.25 & 91.57 & 95.78 &  13.16 & 74.42 & 91.89 & 97.25 &  9.52 & 74.60 & 92.05  \\
              & 100th & 74.12 & 91.52 & 96.04 &  14.28 & 74.29 & 91.90 & 97.38 &  10.20 & 74.62 & 92.01  \\ \midrule
              \multirow{2}{*}{100} &
                  1st & 73.10 & 90.19 & 93.80 &  8.93 & 73.52 & 90.75 & 95.21 &  7.30 & 73.82 & 91.37  \\
              & 100th & 73.08 & 90.05 & 92.70 &  9.34 & 73.45 & 90.62 & 95.36 &  7.41 & 73.90 & 91.29  \\ \midrule
              \multirow{2}{*}{1,000} &
                  1st & 68.90 & 86.85 & 89.51 &  7.09 & 70.68 & 88.50 & 92.67 &  5.40 & 72.80 & 90.51   \\
              & 100th & 68.65 & 86.58 & 90.18 &  7.25 & 70.72 & 88.38 & 92.88 &  5.62 & 72.91 & 90.34   \\ \bottomrule
        \end{tabular}
\end{table*}

\subsubsection{Unlearning DNN on Purchase}
Table~\ref{tbl:purchase-dnn} shows the results on Purchase using DNN. \change{We study the influence of unlearning different data numbers.} 
When unlearning 0.01\% data, \approach reaches 97.15\% accuracy (approaching 97.21\% in naive), 99.70\% consistency, and 33.3$\times$ speed-up compared to the naive method. 
As unlearning more data, the retrain cost rises, and the model accuracy decreases, but it is still close to the naive method ($<$2\%). 
When unlearning $\leq$10\% data, we achieve $>$93\% accuracy and $>$98.7\% consistency with $<$8\% time cost.

\begin{remark}
	\approach works effectively on the multivariate dataset Purchase when unlearning different percentages of training data. It still raises more than 13$\times$ speedup and achieves a high accuracy ($>$93\%) and consistency ($>$98.7\%) when deleting $\leq$10\% data. 
\end{remark}

\subsubsection{Unlearning ResNet-50 on ImageNet} 
Table~\ref{tbl:imagenet-resnet50} shows the results on the ImageNet dataset using a ResNet-50. 
\change{
We also study the performance under different positions and unlearned data numbers, and compare it with the naive method.
When unlearning only 1 data point from the 1st position, \approach reaches 74.42\% top-1 accuracy, and 91.89\% top-5 accuracy, approaching 74.60\% top-1 and 92.05\% top-5 in naive, with 9.5$\times$ speed-up.
When unlearning 100 data from the 1st position, we still get 73.52\% top-1 accuracy, and 90.75\% top-5 accuracy, approaching 73.82\% top-1 and 91.37\% top-5 in naive, with 7.3$\times$ speed-up.
Unlearning 1,000 data continues to decrease the accuracy, however in practical scenes, requests to delete a small amount of data are more common. 
Deleting the front position (1st) brings higher retrain cost, but this influence is very weak. 
A smaller $\varepsilon$ value still improves both top-1 and top-5 accuracy, the consistency, and \revise{decreases the speed-up}. 
}  

\begin{remark}
	\approach can apply to a more complex ImageNet dataset and different unlearning scenes. Section~\ref{sec:eval:effective} has set up many scenarios, such as different $B$ values, $\varepsilon$ values, unlearned positions, unlearned data numbers, and multi-blocks deletion. 
\end{remark}

\begin{remark}
    \change{From simple models (\eg LeNet, DNN) and datasets (\eg MNIST, Purchase) to complex models (\eg VGG, ResNet) and datasets (\eg ImageNet), when unlearning one data, the speed-up decreases from 75$\times$ to 14$\times$, and the consistency decreases from $>99\%$ to 96\%, but the accuracy still approaches naive method with $<1\%$ gaps.}
\end{remark}

\begin{table*}[t]
\caption{Verification of \approach using backdoor triggers. ``Backdoor Position'' is the block with backdoor data. \change{We inject 500 backdoor data with a trigger of 5$\times$5 pixel resolution on MNIST, CIFAR-10, and SVHN ($\approx$1/40 size of the normal image). We inject \revise{1,000} backdoor data with a trigger of 20$\times$20 pixel resolution on ImageNet ($\approx$1/125 size of the normal image), because its images have more pixels and are harder to attack. \revise{Backdoor data is performed on 10 randomly selected classes.} ``Succ'' is the success rate of backdoor data with injected triggers, which verifies the unlearning effect. ``Original Model'' is trained on all data, including backdoor data. ``Model of Naive Unlearning'' is trained on all data except backdoor data, which is unlearned.} ``Acc.'' is the accuracy on the test dataset, and it is top-1 accuracy on ImageNet.
 }\label{tbl:mnist-backdoor}
\centering
\scriptsize
        \begin{tabular}{cccccccccccc} \toprule
             \change{\multirow{2}{*}{Dataset}} & \multirow{2}{*}{\revise{$B$}} & Number of & Backdoor & \multicolumn{2}{c}{\approach $\varepsilon=0.1$ } & \multicolumn{2}{c}{\approach $\varepsilon=0.05$ } & \multicolumn{2}{c}{Model of Naive Unlearning} & \multicolumn{2}{c}{Original Model} \\ \cline{5-12}
             &  & Unlearned Data & Position &  Succ.(\%) & Acc.(\%)  & Succ.(\%) & Acc.(\%)  & Succ.(\%) & Acc.(\%) & Succ.(\%) & Acc.(\%) \\ \toprule
             \multirow{3}{*}{MNIST} & \multirow{3}{*}{100} & \multirow{3}{*}{500}
                &  1st  & 13.0 & 96.25  & 12.6 & 96.36  & 11.2 & 98.07 & 86.6 & 99.02 \\ 
             & & & 20th & 13.4 & 96.33  & 13.0 & 96.38  & 11.6 & 98.13 & 86.7 & 99.05  \\ 
             & & & 60th & 13.3 & 96.33  & 13.3 & 96.35  & 11.6 & 98.12 & 86.9 & 99.01 \\ \midrule
             \multirow{3}{*}{SVHN} & \multirow{3}{*}{1,000} & \multirow{3}{*}{500}
                 &  1st  & 13.0 & 94.45  & 12.8 & 94.63  & 11.0 & 94.68 & 82.5 & 95.10 \\ 
             & &  & 10th & 13.4 & 94.58  & 13.5 & 94.75  & 11.4 & 94.70 & 82.9 & 95.18 \\ 
             & & & 100th & 13.5 & 94.54  & 13.2 & 94.72  & 10.8 & 94.75 & 83.2 & 95.12 \\ \midrule
            \multirow{3}{*}{CIFAR-10} & \multirow{3}{*}{100} & \multirow{3}{*}{500}
                  & 1st  & 14.8 & 90.12  & 13.9 & 90.45  & 11.5 & 91.15 & 79.8 & 91.31 \\ 
             & &  & 10th & 14.6 & 90.35  & 14.2 & 90.57  & 11.6 & 91.20 & 80.5 & 91.30 \\ 
             & &  & 60th & 14.5 & 90.44  & 14.0 & 90.70  & 11.8 & 91.17 & 80.9 & 91.35 \\ \midrule
             \multirow{3}{*}{\revise{ImageNet}} & \multirow{3}{*}{1,000} & \multirow{3}{*}{1,000}
                  & 1st   & 12.3 & 72.69  & 11.8 & 73.11  & 10.9 & 73.80 & 77.8 & 74.51 \\ 
             & &  & 10th  & 11.8 & 72.95  & 11.4 & 73.25  & 11.1 & 73.69 & 78.4 & 74.38 \\ 
             & & & 100th  & 11.6 & 73.16  & 11.2 & 73.37  & 11.0 & 73.72 & 79.5 & 74.20 \\ \bottomrule
        \end{tabular}
\end{table*}

\subsection{Backdoor-based Unlearning Verification}\label{sec:eval:verification}
Verifying whether the requested data is unlearned from model is not an easy task. 
Dishonest model providers may pretend to unlearn data as requested but actually not. 
It is also intractable to prove the completion of unlearning for users, especially on a huge dataset. The removal of a small portion of data can only exert negligible influence to the model. \change{A good verification function $f$ is needed to distinguish $F(D)$ and $U(x_d; F(D))$.} 
\revise{For deep learning models, unlearning small amounts of data has little influence on the model's functionality.} Even if the unlearned data is indeed removed, the model also has a great chance to predict it correctly, because other users may have provided similar data. So it is not practical to verify the unlearning operation with the accuracy of unlearned data. 
Therefore, we resort to the backdoor-based verification method in~\cite{u7-DBLP:journals/corr/abs-2003-04247}. 
\change{We design a crafted trigger and implant it into unlearned data (also called backdoor data), which hardly affects model accuracy. The backdoor data can attack the original model with a high success rate, but hardly attacks the naive unlearned model, which has never seen it. If the backdoor data also hardly attacks \approach, it proves that our method has deleted the unlearned data with backdoor.}
Certainly, our focus here is not to discuss the attack feasibility of this verification in real world, but to prove that \approach can make a successful unlearning.

In Table~\ref{tbl:mnist-backdoor}, we implement verification experiments using backdoor data on four datasets. 
Three different backdoor positions are selected to prove that it is not affected by deletion positions. Model accuracy (``Acc.'') guarantees that \approach hardly reduces the model performance.

On MNIST, the backdoor data only reaches a maximum attack success rate of 13.4\% on \approach, approaching 11.6\% on the naive unlearned model, but away from 86.7\% on the original model. The success rate gap of different positions is $<1\%$. 
On SVHN, \approach reaches 13.3\% average success rate at $\varepsilon=0.1$ and 13.2\% at $\varepsilon=0.05$. The average success rate is 11.1\% on the naive unlearned model and 82.87\% on the original model. \approach only decreases less than 1\% accuracy.
On CIFAR10, the backdoor data has 80.4\% average success rate on the original model. However, the average success rate decreases to 11.63\% after naive unlearning, and 14.33\% after \approach, proving that our method has deleted the backdoor data.
On ImageNet, the backdoor data reaches 77.8\%$\sim$79.5\% success rates on the original model. After naive unlearning, the attack success rate drops to 11.0\%. \approach also reaches similar low success rates, from 11.2\% to 12.3\%. 
The attack success rate can clearly distinguish between the original model and \approach.

\begin{remark}
	\approach can pass the backdoor-based unlearning verification on four datasets. \change{\approach has almost deleted the unlearned data, with hardly decreasing the model performance.} 
\end{remark}

\begin{table}[t]
	\caption{Comparison of \approach, SISA, and the naive method. Only 1 data point is unlearned in each request. $\varepsilon=0.1$ in \approach. SISA has $S$ shards and each shard has $R$ slices. ``Acc.'' is the accuracy on test dataset. ``Speed-up'' is the accelerating effect of retraining, comparing with ``Naive'' (1$\times$). ``PC.'' is the prediction cost.  ``SC.'' is the storage cost.} 
	\label{tbl:compare-naive-sisa}  
	\centering
	\scriptsize
	\begin{tabular}{cccccc} \toprule
		Dataset & Method & Acc. & Speed-up & PC. & SC.  \\ \toprule
		\multirow{3}{*}{MNIST}&Naive & 99.1\% & 1$\times$ & 1$\times$ & 1$\times$   \\  
		&SISA($S$=50,$R$=20) & 96.2\% & 73$\times$ & 50$\times$ & 1,000$\times$ \\   
		&\revise{Ours($B=1,000$)} & \textbf{98.9\%} & \textbf{67$\times$} & 1$\times$ & 1,000$\times$  \\ \midrule 
		\multirow{3}{*}{SVHN}&Naive & 95.3\% & 1$\times$ & 1$\times$ & 1$\times$   \\  
		&           SISA($S$=50,$R$=20) & 90.4\% & 73$\times$ & 50$\times$ & 1,000$\times$  \\  
		&Ours($B=1,000$)& \textbf{94.9\%} & \textbf{75$\times$} & 1$\times$ & 1,000$\times$  \\ \midrule
		\multirow{3}{*}{Purchase}&Naive & 97.2\% & 1$\times$ & 1$\times$ & 1$\times$   \\  
		&           SISA($S$=20,$R$=10) & 95.3\% & 29$\times$ & 20$\times$ & 200$\times$  \\  
		& Ours($B=200$) & \textbf{97.0\%} & \textbf{35$\times$} & 1$\times$ & 200$\times$  \\ \midrule
		\multirow{3}{*}{ImageNet}&Naive & 74.6\% & 1$\times$ & 1$\times$ & 1$\times$   \\  
		&           SISA($S$=10,$R$=20) & 59.5\% & 15$\times$ & 10$\times$ & 200$\times$  \\  
		& Ours($B=200$) & \textbf{74.1\%} & \textbf{14$\times$} & 1$\times$ & 200$\times$  \\ \bottomrule
	\end{tabular}
	\vspace{-3mm}
\end{table}

\subsection{Comparison with Other Methods}

Table~\ref{tbl:compare-naive-sisa} compares \approach with SISA~\cite{u3-DBLP:journals/corr/abs-1912-03817} and the naive method under unlearning only 1 data point. \revise{$S$ is the number of shards and $R$ is the number of slices in SISA. In this comparison, we consider computation time with no parallelization. For model storage, \approach needs to store $B\times$ normal models and SISA needs $SR\times$. Here we choose $B=SR$ to compare them under the same storage overhead.}

\change{
Compared to SISA, \approach can reach a higher accuracy and a similar speed-up under the same storage cost. 
On MNIST, our method can achieve 98.9\% accuracy, very close to 99.1\% in naive unlearning, improving 2.7\% than SISA. Furthermore, \change{SISA needs 50$\times$ prediction time cost while \approach only needs 1$\times$.} The acceleration effect is almost the same. Our method only costs 8.9\% more retrain time than SISA.
On SVHN, our method achieves 94.9\% accuracy, approaching 95.3\% in naive method, while SISA only reaches 90.4\% accuracy. We also get a larger speed-up (75$\times$) than SISA (73$\times$). 
On Purchase, our method improves 1.7\% accuracy and achieves 1.2$\times$ speed-up than SISA. On ImageNet, our method improves 14.6\% accuracy and a similar speed-up. } 

\change{
Compared to the naive method, \approach can accelerate the unlearning process 67$\times$, 75$\times$, 35$\times$, and 14$\times$ on MNIST, SVHN, Purchase, and ImageNet with a similar accuracy (lowering $\leq$0.5\%). Although we need higher storage overhead, at present, the hard disk storage is very cheap, which we discuss in detail in Section~\ref{sec:discuz}. 
}

\revise{The block in our method and the shard in SISA both divide an entire dataset into several disjoint small datasets. But the difference is that SISA uses each shard as an independent dataset to train a model separately, and the final model is just an aggregation of these small models trained on each shard. This will cause a weak learner problem. However, our method trains our model on all blocks, and finally form a complete and powerful unlearned model.}  

\revise{The slice in SISA only considers where we start a retraining as mentioned in Section~\ref{sec:over:symbo}, which is the simple part and only has limited speed-up. On average, it has $\frac{3R}{2R+1}\times$ ($<1.5\times$) speed-up. 
While our method also considers the more important and difficult part, namely where we terminate the retraining.}  

\revise{In addition, the isolation in SISA may degrade the generalization ability of the aggregation model. Its effect is closely related to how the dataset is divided into shards. Poorly dividing will seriously affect the performance of aggregation.}  
\revise{In order to improve the acceleration effect, SISA needs to know which data points have larger probability to be unlearned. This is difficult to achieve in normal unlearning operations. Our method does not need this information. Each data point should have the right to be deleted.} 

\revise{In general, \approach does not need the probabilities of different data being unlearned, and there is no problem of a weak learner. We can delete any number of data from any position. More models in SISA lead to more hyperparameters and prediction cost. }

\begin{remark}
    \change{Compared to SISA, \approach can reach a much higher accuracy and a similar unlearning speed-up under the same storage overhead. \approach also needs no additional prediction cost.  }
\end{remark}

	\section{Discussion}\label{sec:discuz}

\noindent\textbf{Cost Analysis}. We analyze the costs incurred by \approach through model training, use, and maintenance. 
In the phase of model training, our approach makes no differences with a normal training process except storing intermediate models. It is concluded that \approach does not produce additional time overhead, but needs $B\times$ storage occupancy with $B$ stored models. \change{Fortunately, the hard disk storage is cheap now, and 1,000 VGG-11 models only need about 1TB hard disk storage. Besides, this is a one-off expense, and we can use the compression technology on DNNs~\cite{space-compress} to reduce it.} \revise{The storage overhead is linear in the number of $B$. It is completely acceptable comparing with the reduction of retraining overhead.}

When the model is on the shelf, our method needs 1$\times$ prediction time, the same as the original model. However, SISA~\cite{u3-DBLP:journals/corr/abs-1912-03817} must pay out $S\times$ computation for $S$ shards. 
Although one-time-prediction cost is far less than the training cost, it is a continuous burden for model providers. The cost will rise significantly with more predictions.
As for maintenance, model providers can unlearn or incrementally learn some data. 
In an unlearning task, \approach can save much retraining cost by reusing stored models and model stitching. 
In incremental learning, \approach does not destroy the normal model training pattern. Therefore, our method has no additional overhead for incremental learning. Since~\cite{u1-DBLP:conf/sp/CaoY15} transforms training data into summations, it need to update most summation results when adding a new sample, bringing extra cost.

\noindent\textbf{The Choice of $B$}. \revise{$B$ depends on factors such as the training dataset size and the requirement for model accuracy. For a larger dataset (\eg SVHN, ImageNet), we can choose a larger $B$ (\eg 1,000). Increasing $B$ can improve the retraining speed-up with some loss of accuracy.}

\noindent\textbf{Influential Factors for $\Delta$ Curve}.
To further demystify how $\Delta$ curves come and what else factors are dependent, we conduct more experiments to infer these factors such as the distribution, position, and amount of unlearned data. 
The experiments are conducted on the MNIST dataset, and results show there are only slight differences in the curves when deleting data of varying labels; the position and amount do not affect the curvature of $\Delta$ at all. We intend to explore more factors with more datasets in our future work.

\noindent\textbf{Resilience to attacks}. 
It is well known that DLMs are susceptible to adversarial attack, model extraction attack, model inversion attack, poisoning attack (including backdoor attack)~\cite{survey,usenix2021drmi}.
As described in Section~\ref{sec:approach}, our approach merely records the intermediate models during training, but does not make any changes to the trained model. 
\revise{\approach hardly brings additional security risks, but it requires the model service provider to protect stored intermediate models as strictly as the final model.}
As concluded from Section~\ref{sec:eval:verification}, \approach can be used to remove the influence of poisonous data, thus resisting poisoning attack to some extent. Additionally, we envision that stored models are helpful in detecting other attacks which will be further researched in future. 

\noindent\textbf{Differential privacy} \cite{u33-DP-Chaudhuri-2011} provides a privacy guarantee for any individual record in a database. Some recent ideas~\cite{u31-DP-Dwork-2014,u32-DP-Abadi-2016,u34-DP-Jayaraman-2019} apply differential privacy (DP) to machine learning to ensure that model parameters cannot leak the private information of any training data. 
\revise{However, DP and machine unlearning are two totally different technologies.
DP ensures the contribution of each training sample to the target model has a safe range, but the contribution cannot be limited to zero. Otherwise, the model would not learn anything. While machine unlearning requires the contribution of a specific training point to the model is zero. Therefore, the rough application of DP cannot solve the problem of machine unlearning. }

	\section{Related Work}\label{sec:related}
In this section, we review two types of machine unlearning techniques: \emph{parameters manipulation} and \emph{dataset reorganizing}. Then we present the verification of machine unlearning, \change{and the data influence and memorization approaches.}

\noindent\textbf{Parameters Manipulation.} Model providers directly change model parameters to offset influence. 
Tsai \etal \cite{u9-DBLP:conf/kdd/TsaiLL14} proposed a warm start strategy for decremental learning in linear classifiers.
Baumhauer \etal~\cite{u5-DBLP:journals/corr/abs-2002-02730} used a linear filtration method to remove the influence on parameters. But it is specific to certain models and use cases. 
Schelter \cite{u11-DBLP:conf/cidr/Schelter20} introduced a decremental update method to forget data without revisiting training data, only for non-DLMs.
Liu \etal~\cite{u6-DBLP:journals/corr/abs-2003-10933} studied unlearning specific to federated learning. Each user has a trainable dummy gradient generator to eliminate the memorization. It requires the client to do operations.
Wu \etal~\cite{u13-ICML-DeltaGrad-Wu} performed deletions by differentiating the optimization path with Quasi-Newton method. They need cache model parameters and gradients for each iteration.
Golatkar \etal~\cite{u14-CVPR-Golatkar} utilized the fisher information matrix to compute the optimal noise to destroy the information of forgotten data.
\emph{\approach also removes the influence of unlearned data, which is quantified by temporal residual memory. The part of removing impact is then left to model retraining, which reduces the deviation from directly calculating influence.}

\noindent\textbf{Dataset Reorganizing.} It refers to methods of reorganizing the training data. 
Cao \etal~\cite{u1-DBLP:conf/sp/CaoY15} used a statistical query learning to transform training data into summation forms. They only update summations when deleting data. The qualities of transformation functions affect the results.
Ginart \etal~\cite{u2-DBLP:conf/nips/GinartGVZ19} studied the data deletion in K-means. They adopted a tree structure, and divided the large dataset into small subsets layer by layer. Each leaf is an independent K-means instance.  
Bourtoule \etal~\cite{u3-DBLP:journals/corr/abs-1912-03817} divided training data into isolated shards, and trained isolated models on each shard separately. It modifies the original model training pattern. This separation decreases model accuracy because collective weaker models cannot rival a complete model built on the entire dataset.
\emph{Differently, \approach retains the conventional training processing to the largest extent, and thus better guarantees models' performance.}

\noindent\textbf{Verification of Machine Unlearning.} 
To verify a successful unlearning, Sommer \etal~\cite{u7-DBLP:journals/corr/abs-2003-04247} proposed a probabilistic verification method based on backdoor attack. Each user inserts backdoor triggers into his data, as an indicator of unlearning data. If the user data is not deleted, the model outputs the user-specified pollution label for polluted data, otherwise the normal label.
Guo \etal~\cite{u4-DBLP:journals/corr/abs-1911-03030} defined the certified removal for linear classifiers by limiting the max-divergence between the unlearned model and baseline unlearning.
Garg \etal~\cite{u10-DBLP:conf/eurocrypt/GargGV20} analyzed what is expected from a model under a request of deleting data. They used tools from cryptography to explain the right to be forgotten.

\noindent\textbf{Data Influence and Memorization.} 
Koh and Liang \cite{u27-influence2017icml} \revise{measured the influence on a model's prediction w/ or w/o a training point.} They retrained the dataset with a data being discarded, and observed the changes of loss. \textit{While we use updated models to measure whether a data point is completely unlearned.} 
Arpit \etal~\cite{u30-memorization-ICML-noise} found that DNNs memorized both real and noise data. Dropout can degrade the memorization on noise data, without compromising generalization on real data. 
Carlini \etal~\cite{u26-unintended-memorization-D} showed that unintended memorization is a persistent problem in DLMs and proposed a method to evaluate and test it. \textit{While \approach eliminates not only unintended memorization, but also intended memorized information of unlearned data.}

	\section{Conclusion}\label{sec:concl}

To mitigate the urgent requirements of machine unlearning and current technical deficiencies, we propose \approach to accomplish a simple, fast yet cost-effective unlearning for target data. 
Our starting point is to exchange space for time by storing and reusing intermediate models, while ensuring high model accuracy. 
It only retrains the part where residual memory of unlearned data resides, and directly stitch the rest stored models.
Through the evaluation on five datasets and deep models, \approach proves to be superior in reducing unlearning costs and preserving models' performance simultaneously. 
In addition, the unlearning effect is successfully verified via a backdoor attack, and our approach applies to a variety of deep learning models without any interference made to the model.
	\newpage
	
	
	
	
	\bibliographystyle{ACM-Reference-Format}

\end{document}